\documentclass[acmsmall]{acmart}
\usepackage{ragged2e}
\usepackage{booktabs,makecell, multirow, tabularx}
\usepackage{amsmath}
\usepackage{wrapfig}  

\usepackage{subfloat}
\usepackage{subfig}

\AtBeginDocument{%
  }

\acmJournal{JACM}
\acmVolume{37}
\acmNumber{4}
\acmArticle{111}
\acmMonth{8}

\renewcommand\footnotetextcopyrightpermission[1]{}
\begin{document}

\title{A Comprehensive Survey on Multi-modal Conversational Emotion Recognition with Deep Learning}


\author{Yuntao~Shou}
\email{shouyuntao@stu.xjtu.edu.cn}
\affiliation{
  \institution{Central South University of Forestry and Technology}
  \city{ChangSha}
  \state{Hunan}
  \country{China}
  \postcode{410004}
}

\author{Tao Meng}
\authornote{Corresponding Author}
\email{mengtao@hnu.edu.cn}
\affiliation{
	\institution{Central South University of Forestry and Technology}
	\city{ChangSha}
	\state{Hunan}
	\country{China}
    \postcode{410004}
}

\author{Wei~Ai}
\email{aiwei@hnu.edu.cn}
\affiliation{
	\institution{Central South University of Forestry and Technology}
	\city{ChangSha}
	\state{Hunan}
	\country{China}
    \postcode{410004}
}

\author{Fangze Fu}
\email{fufangze@csuft.edu.cn}
\affiliation{
	\institution{Central South University of Forestry and Technology}
	\city{ChangSha}
	\state{Hunan}
	\country{China}
	\postcode{410004}
}
	
\author{Nan Yin}
\email{nan.yin@mbzuai.ac.ae}
\affiliation{
	\institution{Mohamed bin Zayed University of Artificial Intelligence}
	\city{Masdar City}
	\state{Abu Dhabi}
	\country{UAE}
    \postcode{100701}
}

\author{Keqin~Li}
\email{lik@newpaltz.edu}
\affiliation{%
	\institution{State University of New York}
	\city{New Paltz}
	\state{New York}
	\country{USA}
	\postcode{12561}
}

\renewcommand{\shortauthors}{Shou et al.}

\begin{abstract}
Multi-modal conversation emotion recognition (MCER) aims to recognize and track the speaker's emotional state using text, speech, and visual information. Compared with traditional single-utterance multi-modal emotion recognition or single-modal conversation emotion recognition, MCER is more challenging. It requires modeling complex emotional interactions and learning consistent and complementary semantics across multiple modalities. Although many deep learning-based approaches have been proposed for MCER, there is still a lack of systematic reviews summarizing existing modeling methods. Therefore, a timely and comprehensive overview of MCER's recent advances in deep learning is of great significance. In this survey, we provide a comprehensive overview of MCER modeling methods and roughly divide MCER methods into four categories, i.e., context-free modeling, sequential context modeling, speaker-differentiated modeling, and speaker-relationship modeling. Unlike conventional taxonomies based on modality combinations or task-stage decomposition, our framework focuses on how models structurally capture conversational dynamics, speaker roles, and emotional dependencies. In addition, we further discuss MCER's publicly available popular datasets, multi-modal feature extraction methods, application areas, existing challenges, and future development directions. We hope this review provides valuable insights into the current state of MCER research and inspires the development of more effective models.
\end{abstract}




\maketitle

\section{Introduction}
With the development of the mobile Internet, social media has become the main platform for people to communicate with each other \cite{park2016multimodal}. Users can fully express their emotions through multi-modal data such as text, voice, image, and video. Building a multi-modal conversational emotion recognition model using multi-modal data is of vital practical significance for understanding users' true emotional intentions \cite{ghosh2021depression}. Therefore, researchers have been trying to give machines the ability to understand emotions in recent years {\cite{li2022bieru, zhu2021topic, zhang2023m3gat}.}

\begin{wrapfigure}{r}{0.45\textwidth} 
	\centering 
	\includegraphics[width=0.88\linewidth]{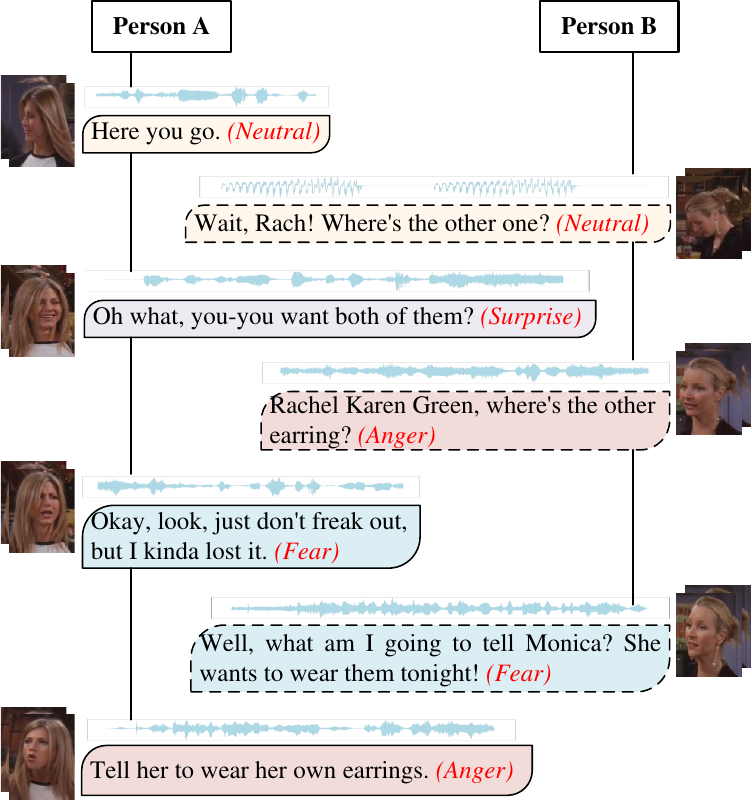}
	\caption{An example of a multimodal conversation emotion recognition dataset which contains three modal features: video, audio, and text. The task of MCER is to identify the emotion label of each speaker at the current moment based on the utterance content (e.g., neutral, angry, surprised, etc.).} 
	\label{fig:multimodal} 
\end{wrapfigure}

\begin{figure*}[htbp]
	\centering
	\subfloat[Example of unimodal vs. multimodal in IEMOCAP.]{\includegraphics[width=0.5\linewidth]{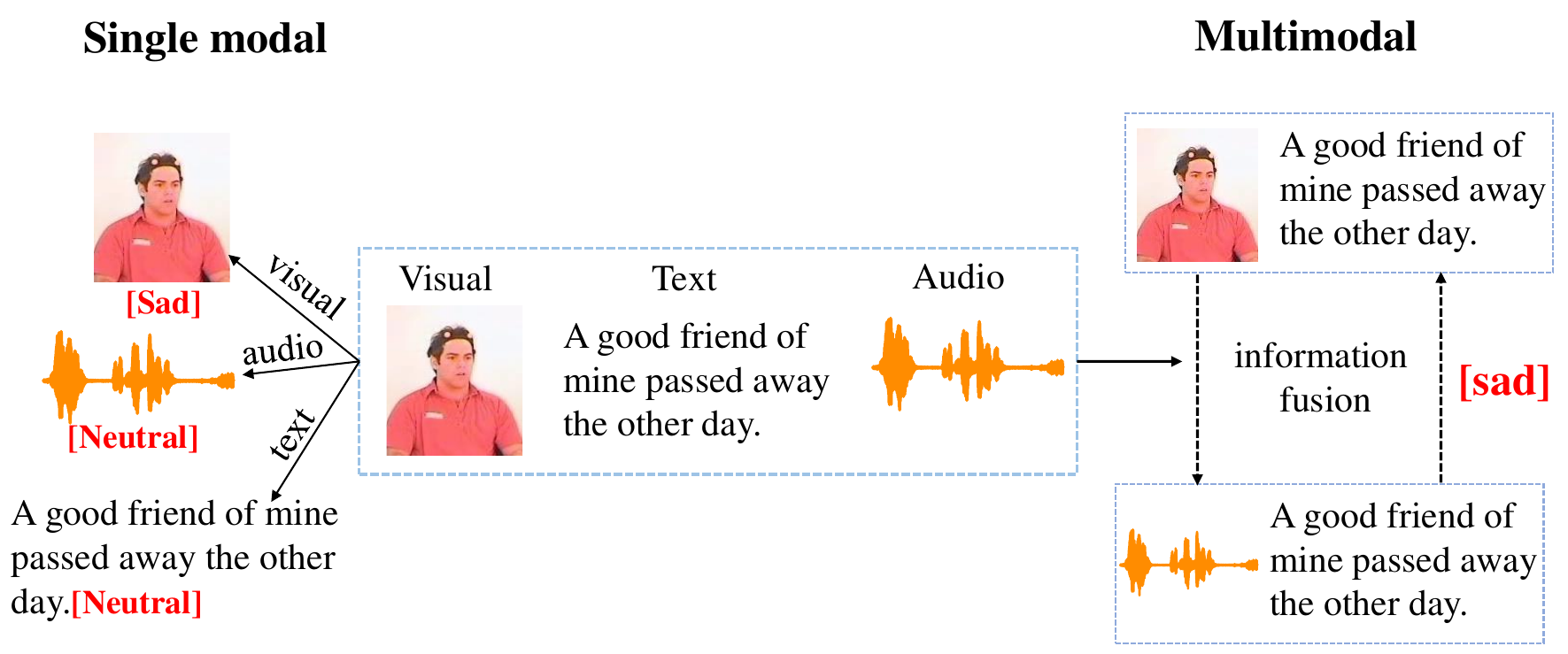}%
		\label{fig:embed_visual_emo_initial_iemocap6}}
	\hfil
	\subfloat[LR-GCN (unimodal)]{\includegraphics[width=0.24\linewidth]{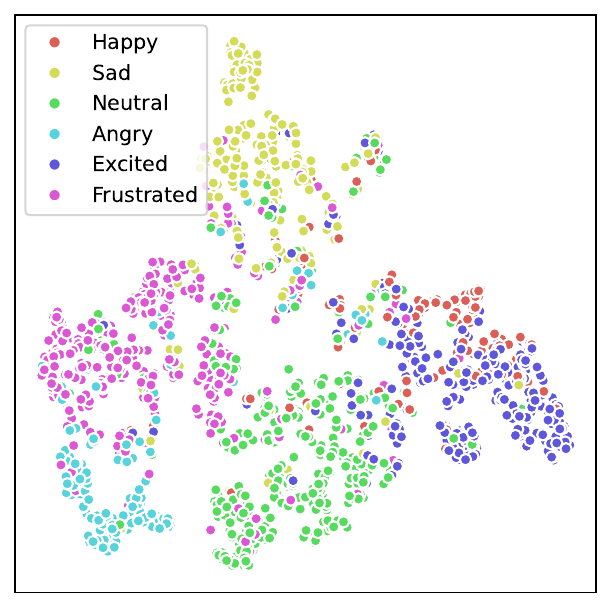}%
		\label{fig:embed_visual_emo_mmgcn_iemocap6}}
	\hfil
	\subfloat[LR-GCN (multimodal)]{\includegraphics[width=0.24\linewidth]{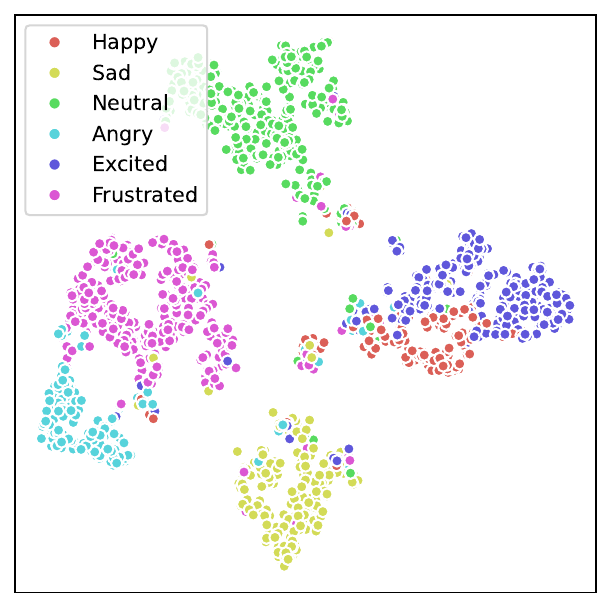}%
		\label{fig:embed_visual_emo_m3net_iemocap6}}
	\hfil
	\caption{{Illustration of the advantage of multimodal fusion in emotion recognition. (a) Example from the IEMOCAP dataset showing that textual modality alone may fail to capture emotional intent (“Neutral”), while audio and visual modalities correctly identify the emotion as “Sad”. (b) Latent space visualization of GS-MCC with unimodal (text-only) input shows overlapping clusters and poor separation between emotion classes. (c) The same visualization under multimodal fusion shows significantly improved class separability, demonstrating the effectiveness of incorporating audio-visual information.}}
	\label{fig:visandexm}
\end{figure*}

Before the emergence of multi-modal conversational emotion recognition, early methods \cite{kim2014convolutional, tzinis2018integrating, zhong2019knowledge, lotfian2019curriculum, wen2024personality} primarily relied on single-modal data, such as text or speech. These approaches mainly focused on modeling contextual dependencies within the same modality and leveraging the semantic content of words or audio signals to recognize emotions \cite{meng2021multi, ying2021prediction, shou2022conversational, ye2024relevance}. However, relying solely on textual information may be insufficient for accurately interpreting a speaker's emotional state, as speakers often express their opinions in a reserved or implicit manner \cite{zhu2023multimodal, ghosal2019dialoguegcn, qin2024language}. For example, a speaker may be veiled in expressing his anger, which may result in a more neutral utterance. In response to the above problems, multi-modal conversational emotion recognition (MCER) technology was proposed to solve the problem of insufficient expression of text semantic information \cite{shou2023czl, shou2023graph, ying2021prediction, qin2024language}. As shown in Fig. \ref{fig:multimodal}, MCER aims to extract semantic information complementary within and between modalities and identify the emotions expressed by speakers in text, audio, and video. One major advantage of MCER is its ability to enhance emotion understanding when the emotional polarity conveyed by text alone is insufficient \cite{lian2023gcnet}. In such cases, the model can leverage visual cues (e.g., facial expressions) and acoustic signals (e.g., tone of voice) to supplement and enrich the emotional representation \cite{10.1145/3503161.3548012, yin2023coco, 10314020, 10113198, wang2024automated}. As illustrated in Fig. \ref{fig:visandexm}(a), the text modality alone predicts the speaker's emotion to be ``Neutral", whereas the audio and visual modalities correctly predicts the speaker's emotion to be ``Sad", highlighting the limitations of relying solely on text in emotional context inference. Furthermore, to validate the effectiveness of multimodal fusion, we provide LR-GCN latent space visualizations for both unimodal and multimodal settings in Fig. \ref{fig:visandexm}(b) and (c). It is evident that the multimodal feature space yields better inter-class separation, especially among subtle emotions such as ``Neutral", ``Frustrated", and ``Sad", demonstrating superior discriminative capability.

However, {unlike traditional single-utterance multimodal emotion recognition or single-modal conversation emotion recognition, MCER is a more challenging issue. It requires consideration of factors such as multimodal context, dialogue scenarios, the speaker's emotional inertia, and the interlocutor's stimulation \cite{chudasama2022m2fnet, zhang2020emotion}.} Powerful deep learning technology \cite{ zhang2024soft} enables MCER to recognize emotion by fusing semantic features with complex emotional interactions. Feature fusion in MCER mainly considers intra-modal contextual semantic information fusion and inter-modal complementary semantic information fusion \cite{huang2020multimodal}. On the one hand, intra-modal contextual semantic information fusion refers to extracting the temporal and spatial dependencies of speaker feature representations in each modality. On the other hand, complementary semantic information fusion between modalities refers to using the interactive information between different modalities to enhance the emotional understanding ability of the model. MCER synergistically improves the effect of emotion recognition by fusing the characteristics of various modal data, which has important theoretical significance for processing and understanding multi-modal data \cite{9905904, 10050091, zhang2023m3gat}.

{Despite the growing number of researchers focusing on new models and methods for multimodal conversation emotion recognition \cite{ghosal2018contextual, chudasama2022m2fnet, liu2022multi}, there is still a lack of understanding regarding the theoretical and methodological classification of multimodal conversational emotion recognition, particularly those based on deep learning.} To the best of our knowledge, this survey is the first comprehensive survey focusing on deep learning in multi-modal conversation emotion recognition. {Existing reviews mainly focus on multimodal emotion recognition with modal combination \cite{zhang2024deep} or multimodal emotion recognition with task stage decomposition (i.e., feature extraction, feature fusion and classification) \cite{zhu2023multimodal}, without fully considering conversational dynamics, speaker roles and emotional dependencies.}

\begin{figure*}
	\centering
	\includegraphics[width=0.7\linewidth]{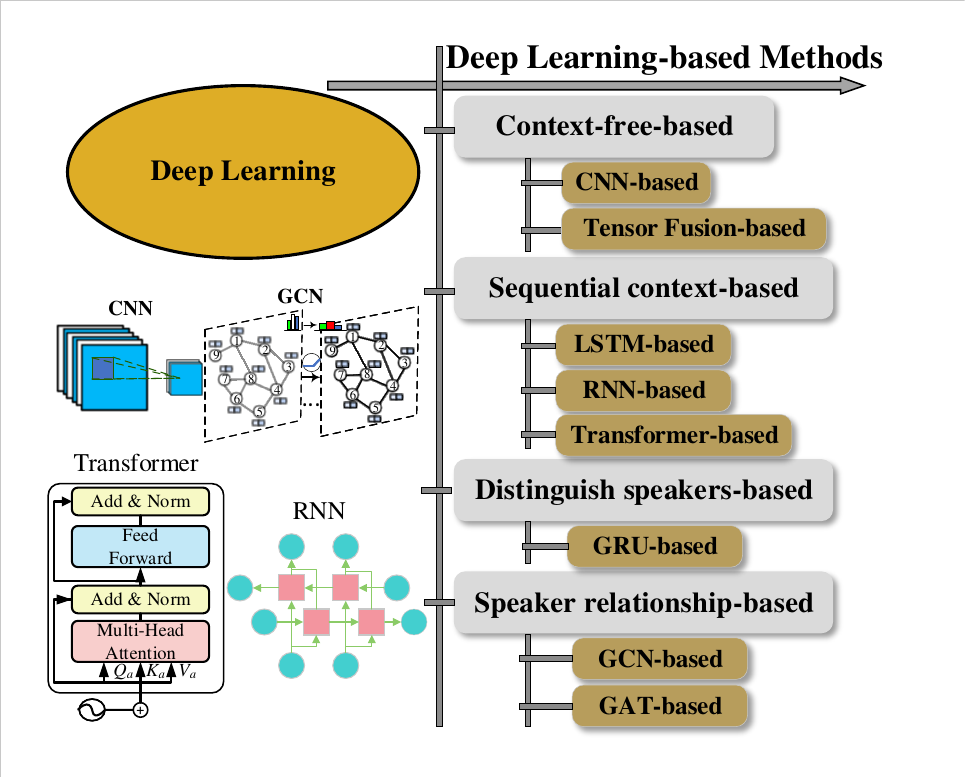}
	\caption{A taxonomy of modeling approaches for multi-modal conversational emotion recognition in conversation. We categorize existing MCER methods into four categories, i.e., context-free modeling, sequential context modeling, distinguishing-speaker modeling, and speaker-relationship modeling.}
	\label{fig:methods}
\end{figure*}

{Different from previous taxonomies, we propose a novel classification framework that emphasizes how methods characterize and model conversational dynamics. Specifically, we categorize MCER methods into four distinct types: context-free modeling, sequential context modeling, speaker-differentiated modeling, and speaker-relationship modeling, as illustrated in Fig. \ref{fig:methods}.}

{Based on the above framework, this survey systematically reviews the key research efforts in MCER. First, we introduce several widely-used and publicly available datasets, along with commonly adopted feature extraction methods across modalities. Next, we detail the proposed modeling taxonomy and comprehensively analyze representative methods within each category. We then discuss evaluation metrics frequently used in MCER experiments. Following that, we examine real-world applications and key challenges faced in this field. Finally, we outline promising directions for future research.}


The contributions made in this paper are summarized as follows:

\begin{itemize}
	\item \textbf{New Taxonomy:} We provide a new taxonomy for multi-modal conversational emotion recognition. Specifically, we classify existing MCER methods into four groups: context-free modeling, sequential context modeling, distinguishing-speaker modeling, and speaker-relationship modeling.
	
	\item \textbf{Comprehensive Review:} This paper provides the most comprehensive review of deep learning and machine learning algorithms for MCER. For each modeling approach, we provide representative models and make corresponding comparisons.
	
	\item \textbf{Abundant Resources:} We collect relevant resources about MCER, including state-of-the-art models and publicly available datasets. This paper can serve as a practical guide for learning and developing different emotion recognition algorithms.
	
	\item \textbf{Future Directions:} We analyzed the limitations of existing MCER methods and proposed possible future research directions in many aspects, such as the collaborative generation of multi-modal data, the deep fusion of multi-modal features, and the unbiased learning of multi-modal emotions.
\end{itemize}

The paper is organized as follows: Section \ref{sec:sec2} summarizes the publicly available and popular datasets in the field of MCER. Section \ref{sec:sec3} illustrates the background, definitions, and commonly used feature extraction techniques for MCER. Section \ref{sec:sec4} broadly divides MCER methods into four categories and analyzes their advantages and disadvantages. Section \ref{sec:sec5} summarizes some commonly used evaluation metrics for MCER tasks. Section \ref{sec:sec6} gives the performance of different algorithms on the IEMOCAP and MELD data sets. Section \ref{sec:sec7} discusses the real-life applications of MCER. Section \ref{sec:sec8} illustrates the problems of existing research and Section \ref{sec:sec9} gives directions for future research. Finally, we conclude the work of this paper.

\section{Popular Benchmark Datasets}
\label{sec:sec2}
Table \ref{table:datasets} presents seven publicly available emotion recognition benchmark datasets. We counted the release time, modality, and open-source URL for each dataset. As shown in Table \ref{tab:data}, we also counted the distribution of the data set on different emotional labels, and the data showed a long-tail distribution.

\begin{table*}[htbp]
	\caption{Publicly available benchmark datasets in multi-modal conversational emotion recognition.}
	\label{table:datasets}
	\centering
	\renewcommand\arraystretch{1}
	\scalebox{0.7}{
	\setlength{\tabcolsep}{3mm}{
		\begin{tabular}{cccc}
			\toprule
			Datasets    & Year & Modality         & Available at                                                        \\ \midrule
			IEMOCAP \cite{busso2008iemocap}     & 2008 & Text,Video,Audio & \href{https://sail.usc.edu/iemocap/}{https://sail.usc.edu/iemocap/}                                       \\
			MELD \cite{poria2019meld}       & 2019 & Text,Video,Audio & \href{https://web.eecs.umich.edu/~mihalcea/downloads/MELD.Raw.tar.gz}{https://web.eecs.umich.edu/$\sim$mihalcea/downloads/MELD.Raw.tar.gz} \\
			DailyDialog \cite{li2017dailydialog} & 2017 & Text             & \href{https://huggingface.co/datasets/daily\_dialog}{https://huggingface.co/datasets/daily\_dialog}                       \\
			EmoryNLP \cite{zahiri2017emotion}   & 2017 & Text             & \href{https://github.com/emorynlp/character-mining}{https://github.com/emorynlp/character-mining}          \\
			SEMAINE \cite{mckeown2011semaine}  &  2012          &   Text,Video,Audio        & \href{https://semaine-db.eu/}{https://semaine-db.eu/}   \\
			EmotionLines \cite{hsu2018emotionlines} &   2018    &    Text      & \href{https://doraemon.iis.sinica.edu.tw/emotionlines/index.html}{https://doraemon.iis.sinica.edu.tw/emotionlines/index.html}   \\
			EmoContext \cite{chatterjee2019understanding}    &    2019    &  Text          & \href{https://www.humanizing-ai.com/emocontext.html}{https://www.humanizing-ai.com/emocontext.html}  \\
			\bottomrule
	\end{tabular}}}
\end{table*}

\begin{table}[htbp]
	\centering
	\caption{Distribution of seven conversational emotion recognition datasets on different emotion labels.}
	\label{tab:data}
	\renewcommand\arraystretch{1}
	\scalebox{0.77}{
	\setlength{\tabcolsep}{1.8mm}{
	\begin{tabular}{cccccccc}
		\toprule
		Labels            & IEMOCAP & MELD  & EmoContext & EmotionLines & EmoryNLP & DailyDialog & SEMAINE \\ \midrule
		Neutral           & 1,708   & 6,436 & -          & 6,530        & 15,104   & 855,72      & -       \\
		Happiness/Joy     & 648     & 2,308 & 4,669      & 1,710        & 11,020   & 12,885      & 93      \\
		Surprise/Powerful & -       & 1,636 & -          & 1,658        & 4,252    & 1,823       & -       \\
		Sadness           & 1,084   & 1,002 & 5,838      & 498          & 3,376    & 1,150       & 58      \\
		Anger/Mad         & 1,103   & 1,607 & 5,954      & 772          & 5,328    & 1,022       & 41      \\
		Disgust           & -       & 361   & -          & 338          & -        & 354         & 7       \\
		Fear/Scared       & -       & 358   & -          & 255          & 6,584    & 74          & 3       \\
		Frustrated        & 1,849   & -     & -          & -            & -        & -           & -       \\
		Excited           & 1,041   & -     & -          & -            & -        & -           & -       \\
		Other             & -       & -     & 21,960     & -            & 4,760    & -           & 197     \\ \bottomrule
	\end{tabular}}}
\end{table}

\subsection{IEMOCAP}
{The interactive emotional dyadic motion capture database (IEMOCAP)} dataset \cite{busso2008iemocap} was released in 2008 and contains 12.46 hours of conversations. The IEMOCAP dataset contains three modal features, i.e., video, audio and text, and it is the first multi-modal dataset for MCER. In the IEMOCAP dataset, ten theater actors express specific emotion categories (i.e., sad, neutral, frustrated, anger, happy, excited) through binary dialogue. To ensure the consistency and accuracy of annotation, each sentence is annotated by multiple experts.

\subsection{MELD}
{The multimodal emotionLines (MELD)} dataset \cite{poria2019meld} is from the classic TV series Friends, which contains text, video and audio data. The MELD dataset contains a total of 13,709 video clips, and each sentence is labeled as a specific emotion (i.e., anger, neutral, fear, disgust, surprise, joy, disgust). In addition, the MELD dataset is also annotated by neutral, negative and positive three-category emotion. To ensure the consistency and accuracy of annotation, each sentence is annotated by multiple experts.

\subsection{DailyDialog}
The DailyDialog dataset \cite{li2017dailydialog} is a multi-turn dialogue dataset about daily chat scenarios, which only contains text modalities. The DailyDialog dataset contain 13, 000 dialogues and labels each sentence with intention (i.e., inform, commissive, directives, questions) and emotion (surprise, sadness, fear, happiness, disgust, anger). Each sentence is annotated jointly by three experts.

\subsection{EmoryNLP}
EmoryNLP \cite{zahiri2017emotion} is a unimodal dataset, containing only text modalities. The EmoryNLP dataset contains 12,606 utterances, and each utterance is annotated with seven emotions: peaceful, scared, crazy, powerful, sad, happy, and neutral. EmoryNLP dataset is divided into training set, testing set and validation set.

\subsection{SEMAINE}
{The sustained emotionally colored machine-human interaction (SEMAINE)} \cite{mckeown2011semaine} is a multi-modal conversation data set, which contains four binary conversations between robots and humans. The SEMAINE data set has 95 dialogues with a total of 5798 sentences. Four emotional dimensions are marked: Valence, Arousal, Expectancy, and Power. Valence, Arousa, and Expectancy are continuous values in the range [-1, 1], and the size of the SEMAINE data set is small.

\subsection{EmotionnLines}
The EmotionLines dataset \cite{hsu2018emotionlines} comes from binary conversations between Friends and Facebook, and it only contains text data. The EmotionLines dataset contains 1,000 dialogues with a total of 29,245 sentences. Seven categories of emotions are marked: neutral, fear, surprise, sadness, anger, happiness, disgust. The EmotionLines dataset is rarely used in conversational emotion recognition.

\subsection{EmoContext}
{The emotion contextual detection (EmoContext)} dataset \cite{chatterjee2019understanding} only contains text data. It has a total of 38,421 dialogues and a total of 115,263 sentences. Three types of emotions are marked: happiness, sadness, and anger. Although the EmoContext data is large, it is rarely used in conversational emotion recognition because it only contains text data.

{\subsection{CH-SIMS}}

{The Chinese single-label multimodal sentiment analysis (CH-SIMS) \cite{yu-etal-2020-ch} is a benchmark dataset designed for Chinese multimodal sentiment analysis tasks. This dataset is constructed from real Chinese video conversations, covering three modalities: text, audio, and vision, and has a good foundation for multimodal fusion and alignment. CH-SIMS contains a total of about 10,000 Chinese sentence-level samples, all of which are accompanied by manually annotated continuous sentiment intensity labels (ranging from -1 to +1) and single-label classifications (positive, neutral, and negative). Compared with mainstream English multimodal sentiment analysis datasets such as IEMOCAP and MELD, CH-SIMS is more in line with the characteristics of Chinese language expression, especially when faced with semantic ambiguity (e.g., irony, sarcasm, etc.), there may be significant inconsistencies between modalities.}

{\subsection{MuSE}}

{Multimodal Sentiment dataset (MuSE) \cite{stappen2021muse} is a multilingual and multimodal emotion recognition dataset, which aims to study the performance of multimodal emotion modeling in different languages and cross-cultural contexts. The dataset contains natural speech videos from English and Spanish, with about 2,800 samples, covering three modalities: text, audio, and vision. Each video is recorded by real participants expressing freely around a specific topic, and provides continuous emotion labels, including valence, arousal, and dominance. The labels are derived from the fusion of self-reports and third-party manual evaluations to enhance the objectivity and consistency of annotations. At the modality level, MuSE provides high-quality speech features, expressions, and body posture information, and is equipped with precisely aligned text transcription data, making it suitable for research tasks such as multimodal fusion, modal alignment analysis, and modal inconsistency modeling. In addition, since the dataset has both multilingual and multimodal characteristics, it provides an important experimental platform for cross-lingual emotion transfer learning, multimodal collaborative modeling, and emotion recognition in low-resource languages.}

\begin{figure}
	\centering
	\includegraphics[width=1\linewidth]{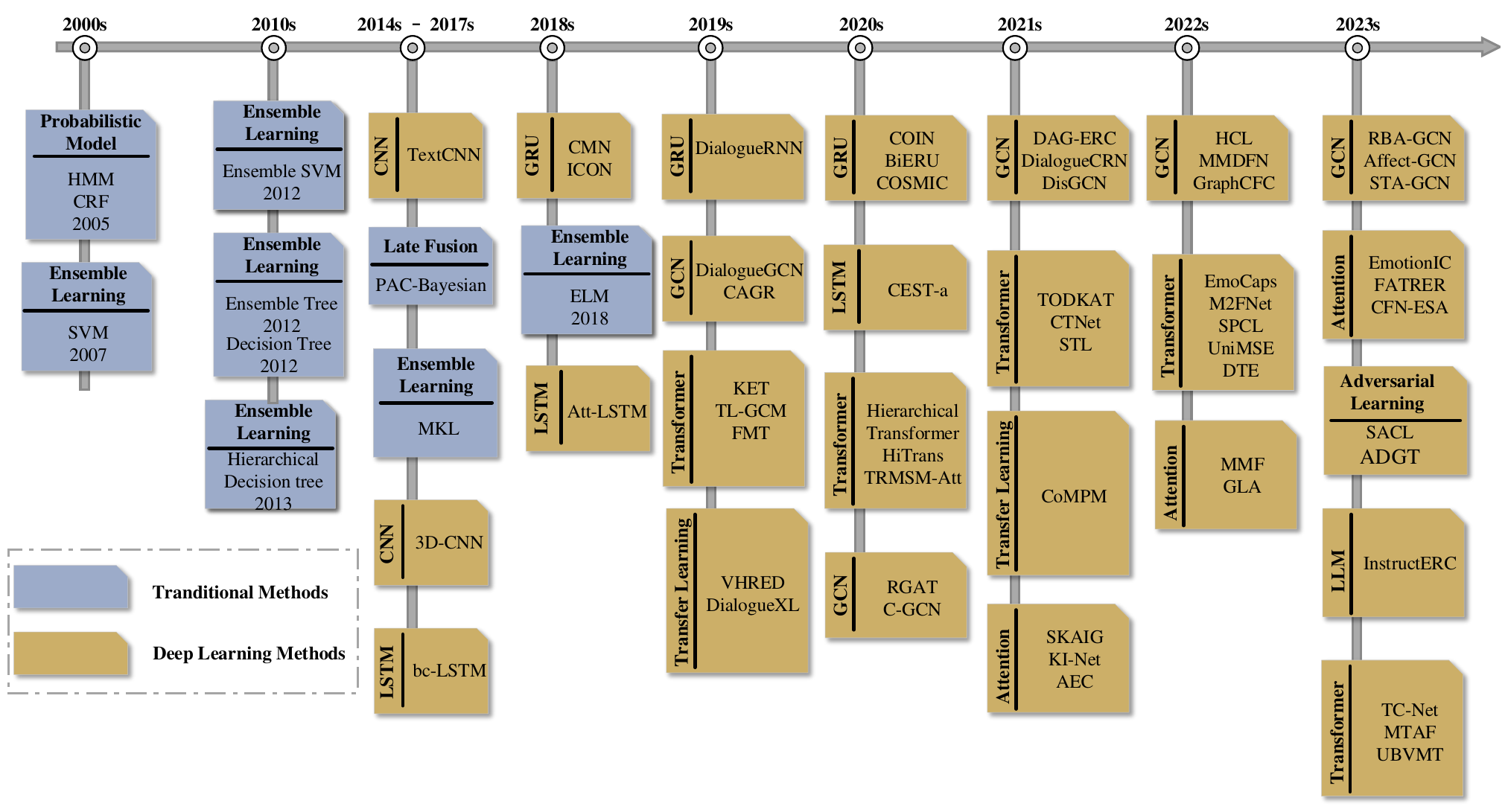}
	\caption{Timeline of multimodal conversational emotion recognition algorithms.}
	\label{fig:tree}
\end{figure}

\section{Background, Definition, and Feature Extraction}
\label{sec:sec3}
\subsection{Background}
As shown in Fig. \ref{fig:tree}, we counted multi-modal conversational emotion recognition algorithms from 2000 to 2023. As can be seen from the figure, before 2018, traditional machine learning algorithms were mainly used, and then deep learning algorithms gradually became the main ones. Next, we briefly return to the main development history of the MCER algorithm.

\subsubsection{Brief History of Conversational Emotion Recognition}
The emotion recognition method based on the dictionary is the earliest used for emotion recognition \cite{hardeniya2016dictionary}, which motivated early work on naive bayes method \cite{frank2000naive}. {With the widespread application of machine learning algorithms in classification tasks, representative algorithms such as the support vector machine (SVM) \cite{rozgic2012ensemble, hu2007gmm} and binary decision tree \cite{cichosz2007emotion, lee2011emotion, liu2018speech} have also gained prominence in emotion recognition.} The above-mentioned method determines the category of emotion by learning the polarity and occurrence frequency of emotional words in the text, which is difficult to extract the semantic information and context information.

Encouraged by the success of {convolutional neural networks (CNNs)} in computer vision (CV), CNNs began to be migrated to text classification tasks and received extensive research attention {\cite{khare2020time, kwon2021mlt, kollias2020exploiting}}. In 2017, Poria et al. \cite{poria2017context} used {long short-term memory (LSTM)} for the first time to resolve dependencies between contexts. Since then, improvements, extensions and applications of LSTMs and {gated recurrent units (GRUs)} have increased {\cite{hazarika2018conversational, majumder2019dialoguernn, hazarika2018icon, rajamani2021novel}}. Until recently, many {graph neural networks (GNN)}-based methods (e.g., {\cite{ghosal2019dialoguegcn, ishiwatari2020relation, shen2021directed, li2023graphcfc, zadeh2018multimodal}}) emerged. Apart from CNNs, {recurrent neural networks (RNNs)}, and GNNs, many alternative Transformer-based methods (e.g., {\cite{zhong2019knowledge, li2020hitrans, zhu2021topic}}) have been developed in the past decades. We detail the categories to which these algorithms belong in Section \ref{sec:sec4}.

\subsubsection{Multi-modal Conversational Emotion Recognition Versus Traditional Machine Learning}
MCER methods based on traditional machine learning {\cite{rozgic2012ensemble, cichosz2007emotion, lee2011emotion, liu2018speech, lin2005speech, hu2007gmm}} are closely related to hand-extracted features, which have attracted increasing attention from the data mining and emotion recognition communities. These methods aim to learn the feature embeddings of raw data for subsequent downstream tasks such as classification and clustering. The classic conversational emotion recognition method based on machine learning is to use support vector machine to map emotional features to a hyperplane and classify them {\cite{rozgic2012ensemble, bhavan2019bagged}}. However, these methods require a large amount of high-quality labeled data.

\subsubsection{Conversational Emotion Recognition Versus Convolutional Neural Network}
The CNN-based emotion recognition methods {\cite{lu2023exploring, kwon2021mlt, kollias2020exploiting}} are the first deep learning method to solve the emotion classification problem historically \cite{kim2014convolutional}. These CNN-based methods employ convolutional filters to extract semantic features of text so that the model can use supervised learning to understand the meaning of text. Similar to machine learning algorithms, CNN can also map emotional features into vector space through mapping functions. The difference is that this mapping function is learned in an end-to-end manner. Since the convolution kernel extracts local receptive field information, CNN cannot contain contextual semantic information.

\subsubsection{Multi-modal Conversational Emotion Recognition Versus Recurrent Neural Network}
The RNN-based emotion recognition methods {\cite{ma2019emotion, tao2018advanced, majumder2019dialoguernn, kollias2020exploiting}} are developed on the basis of CNN, but they believe that contexts should be mutually influential and interdependent {\cite{ma2019emotion, tao2018advanced}}. These RNN-based methods usually use LSTM or GRU (to avoid gradient disappearance or gradient explosion) to extract semantic features including context. Similar to CNN, RNN can also map emotional features into vector space through mapping functions in an end-to-end manner.

\subsubsection{Multi-modal Conversational Emotion Recognition Versus Transformer}
Similar to the RNN-based emotion recognition method, the Transformer-based emotion recognition methods {\cite{huang2020multimodal, lian2021ctnet, tsai2019multimodal, rahman2020integrating}} also extract semantic information including context, and completes subsequent emotion classification based on this \cite{lian2021ctnet}. However, unlike RNN, Transformer's sequential context modeling ability is better than RNN. Therefore, the accuracy of the Transformer-based emotion recognition methods are significantly better than RNNs.

\subsubsection{Multi-modal Conversational Emotion Recognition Versus Graph Neural Network}
The GNN-based emotion recognition methods {\cite{lin2022modeling, zhang2023m3gat, li2023graphcfc, ishiwatari2020relation, shen2021directed}} inherit the idea of the RNN method, i.e., the contexts should interact and depend on each other \cite{ghosal2019dialoguegcn}. On the basis of RNN, GNNs believe that there is also a relationship of mutual influence between speakers. Therefore, GNNs model the dialogue relationship between speakers through the inherent properties of the graph structure.

\begin{table*}[htbp]
	\caption{Some symbols commonly used in the paper.}
	\label{tab:notation}
	\centering
	\renewcommand\arraystretch{1}
    \scalebox{0.86}{
	\setlength{\tabcolsep}{4mm}{
		\begin{tabular}{ll}
			\toprule
			Notations   & Descriptions \\ \midrule
			$\mid\cdotp\mid$  & The length of the set.         \\
			$\bigodot$        & Element-wise product.         \\
			$\mathcal{G}$ & A graph.         \\
			$\mathcal{V}$    & A set of nodes in a graph.         \\
			$v$            & {A node $v \in V$.}  \\
			$\mathcal{E}$  & A set of edges in a graph.   \\
			$e_{ij}$         & An edge $e_{ij} \in E$.       \\
			$N(v)$         & The neighbors of a node $v$.  \\
			$S$            & A speaker.         \\
			$u$            & An utterance.       \\
			$K$            & The context window size. \\
			$M$            & The number of the speakers. \\
			$L$            & The number of utterances in a dialogue. \\
			$U$            & The set of contextual utterence. \\
			$\mathcal{R}$  & The type of edge.  \\
			$\mathcal{W}$  & Learnable parameters. \\
			$\textbf{A}$            & The adjacency matrix of a graph. \\
			$m$            & The node properties of the graph. \\
			$x^t \in \mathbb{R}^d$          &  $d$-dimensional text feature vectors.                           \\
			$x^a \in \mathbb{R}^k$          &   $k$-dimensional audio feature vectors.                          \\
			$x^v \in \mathbb{R}^h$          & $h$-dimensional  video feature vectors.                           \\
			$x$            & Concatenated video, audio and text feature vectors. \\
			\bottomrule
	\end{tabular}}}
\end{table*}

\subsection{Definitions and Preliminaries}
The symbols used in this paper are listed in Table \ref{tab:notation}. Now, we define the sets needed to understand this paper. In particular, we use uppercase letters for matrices and lowercase letters for vectors.

\textit{Definition 1 (Utterances context)}
The multi-modal conversational emotion recognition task aims to recognize the emotional changes (e.g., happiness, and sadness, etc) of speakers $\{S_1, S_2, \ldots, S_M\}$ at the current moment $t$ in a dialogue. $L$ represents the number of utterances in a dialogue, $U$ represent a set of contextual utterances, and $U=\{u_1, u_2, \ldots, u_L\}$.

The MCER task aims to correctly classify each utterance by incorporating contextual information. At the current moment $t$, the model needs to infer the speaker's emotion based on the context information $\{u_1, u_2, \ldots, u_{t-1}\}$. We assume that the context window size is set to $K$. The set of contextual utterances is defined as follows:

\begin{equation}
	C_{\lambda}=\left\{u_i \mid i \in[t-K, t-1], u_i \in U_\lambda,\mid C_{\lambda} \mid \leq K\right\}
\end{equation}

When the context window size is 6, the speaker's contextual utterances and predicted utterances are shown in Table \ref{tab:task}.

\textit{Definition 2 (Dialogue graph)}
A dialogue graph is represented as $\mathcal{G}=\{\mathcal{V}, \mathcal{E}, \mathcal{R}, \mathcal{W}\}$, where $\mathcal{V}$ is a set of nodes in the graph, $\mathcal{E}$ is a set of edges, $v_i \in \mathcal{V}$ represents the $i$-th node, $e_{ij}=(v_i, v_j)\in \mathcal{E}$ represents a directed edge from $v_i$ to $v_j$, the relationship $r_{ij} \in \mathcal{E}$ represents that there is a dialog relationship between nodes $v_i$ and $v_j$. The neighbor nodes of node $v$ are represented as $N(v)=\{u \in \mathcal{V}|(v,u) \in \mathcal{E}\}$. $\textbf{A} \in {\mathbb{R}^{n \times n}}$ means the adjacency matrix with $\textbf{A}_{ij}=1$ if $e_{ij} \in \mathcal{E}$, otherwise $\textbf{A}_{ij}=0$. $\textbf{X} \in \mathbb{R}^{m \times m}$ represents the node properties of the graph. For the MCER task based on GCN, the speaker's utterance information is regarded as the node of the graph, and the dialogue relationship information between speakers is regarded as the edge of the graph.

\textit{Definition 3 (Problem definition)} For a given multi-modal utterance sequence $U$, the MCER task requires using the utterance context information to determine a deep neural network $F\left( {{u_i}} \right)$ so that the output emotion label $\hat{y}_{i}$ is as close as possible to the real emotion label ${y_i}$, $i \in \left\{ {1,...,L} \right\}$. Deep neural networks can solve the optimal parameters by minimizing loss, and its loss is defined as:
\begin{equation}
	\min _{F} \frac{1}{L} \sum_{i=1}^{L} \mathcal{L}\left(\hat{y}_{i}=F\left(u_{i}\right), y_{i}\right)
\end{equation}
where $L$ represents the number of utterances in the dialogue, $\mathcal{L}$ is an indicator function.

\begin{table}[!t]
	\caption{We assume that there are three speakers in a dialogue, and the window size $K$ of the dialogue is set to 6. The dialogue process is as follows:}
	\centering
	\label{tab:task}
	\renewcommand\arraystretch{1}
    \scalebox{0.96}{
	\setlength\tabcolsep{10mm}{
		\begin{tabular}{ccc}
			\toprule
			{Speaker}              & Utterences & Description     \\ \midrule
			$C_a$,$C_b$,$C_c$ & ${{u}_1^a,{u}_3^a},{{u}_2^b,{u}_5^b},{{u}_4^c,{u}_6^c}$ & Contextual utterances\\
			$S_b$    & $u_7^b$     & Predicted utterance                                 \\ \bottomrule
	\end{tabular}}}
\end{table}

\begin{figure*}
	\centering
	\includegraphics[width=1\linewidth]{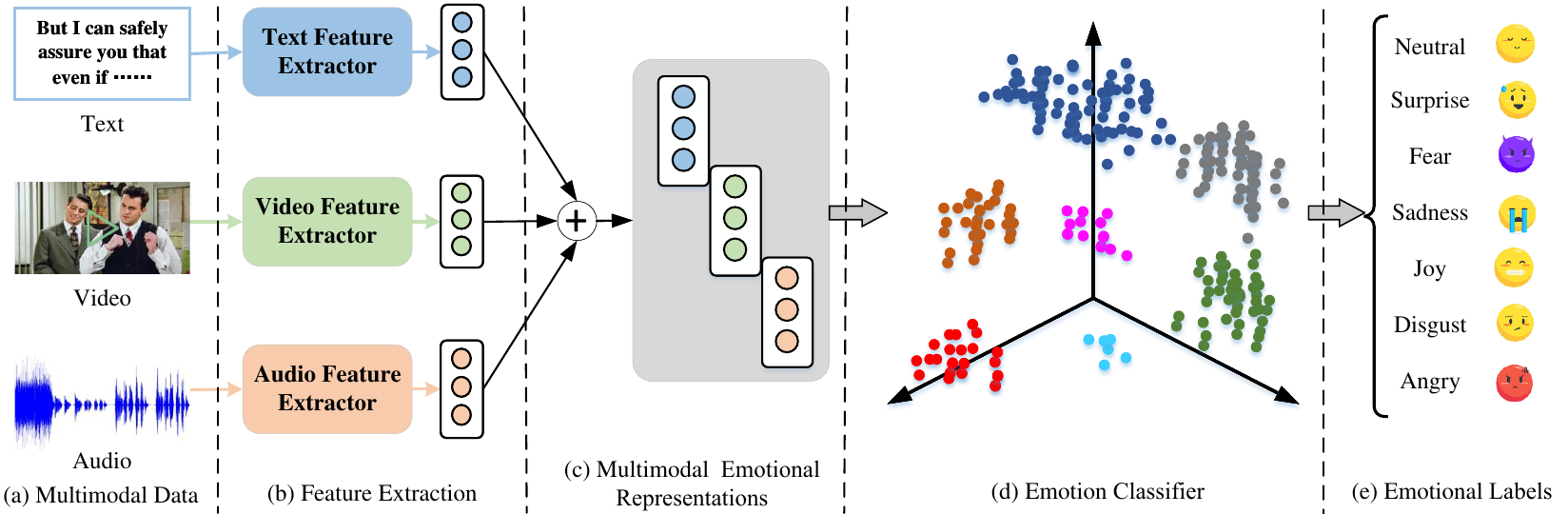}
	\caption{The proposed MCER methods mainly include multi-modal feature extraction, multi-modal emotion representation, and emotion classifier.}
	\label{fig:feature-fusion}
\end{figure*}

From the development history and related preliminary definitions of MCER, it can be seen that the process of multi-modal conversation emotion recognition mainly includes three aspects: multi-modal feature extraction, multi-modal feature fusion representation, and emotion classification. The overall process is shown in Fig. \ref{fig:feature-fusion}, and we will provide a comprehensive overview of these three aspects below.

\subsection{Multi-modal Feature Extraction}
Multi-modal feature extraction (e.g., text, video and audio, etc) is one of the important techniques for emotion analysis. In this section, we introduce the process of using feature extraction methods to perform data preprocessing on text, video, and audio, and list some commonly used feature extraction methods. As shown in Table \ref{tab:feature}, we count the multi-modal feature extraction techniques used by many deep learning methods.

\begin{table*}[htbp]
	\caption{Feature extraction methods for text, video and audio features used by different emotion recognition techniques.}
	\label{tab:feature}
	\centering
	\renewcommand\arraystretch{1}
	\scalebox{0.75}{
		\setlength{\tabcolsep}{1mm}{
			\begin{tabular}{cccc|cccc}
				\toprule
				Methods  & Text            & Video        & Audio     & Methods     & Text     & Video            & Audio     \\ \bottomrule
				THMM \cite{morency2011towards}     & Polarized words & OKAO Vision  & OpenEAR   & CMN  \cite{hazarika2018conversational}       & TEXT-CNN &  3D-CNN       &  openSMILE   \\
				SVM \cite{perez2013utterance}      & Bag-of-words    & CERT         & OpenEAR   & Att-BiLSTM \cite{poria2017context} & TEXT-CNN & 3D-CNN        &  openSMILE   \\
				MKL \cite{poria2015deep}     & Word2vec        & CLM-Z        & openSMILE & ICON  \cite{hazarika2018icon}      & TEXT-CNN &   3D-CNN     &  openSMILE    \\
				SAL-CNN \cite{wang2017select}  & Word2vec        & CLM-Z        & COVAREP   & DialogueRNN \cite{majumder2019dialoguernn} & TEXT-CNN &    3D-CNN     &   openSMILE  \\
				TFN \cite{zadeh2017tensor}     & GLOVE           & Facet        & COVAREP   & DialogueGCN \cite{ghosal2019dialoguegcn}& TEXT-CNN &  3D-CNN       &  openSMILE   \\
				LMF \cite{liu2018efficient}     & GLOVE           & Facet        & COVAREP   & COIN  \cite{zhang2021coin}      & TEXT-CNN &     3D-CNN     &  openSMILE  \\
				HFFN \cite{mai2019divide}    & GLOVE           & Facet        & COVAREP   & CESTa \cite{wang2020contextualized}      & TEXT-CNN &   3D-CNN      &   openSMILE  \\
				LMFN  \cite{mai2019locally}    & GLOVE           & Facet        & COVAREP   & EmoCaps \cite{li2022emocaps}    & BERT     &    3D-CNN     &   openSMILE  \\
				GME-LSTM \cite{mai2019locally} & GLOVE           & Facet        & COVAREP   & MM-DFN \cite{hu2022mm}     & TEXT-CNN &      3D-CNN    &  openSMILE  \\
				MARN \cite{zadeh2018multi}     & GLOVE           & Facet        & COVAREP   & M2FNet  \cite{chudasama2022m2fnet}    & RoBERTa  & Mel Spectrograms & MTCNN     \\
				MFN \cite{zadeh2018memory}     & GLOVE           & Facet        & COVAREP   & GraphCFC \cite{li2023graphcfc}   & TEXT-CNN & openSMILE        & 3D-CNN    \\
				RAVEN \cite{wang2019words}   & GLOVE           & Facet        & COVAREP   & UniMSE \cite{hu2022unimse}     & T5       & openSMILE        & 3D-CNN    \\
				SWRM \cite{wu2022sentiment}    & BERT            & Facet        & COVAREP   & EmotionIC \cite{yingjian2023emotionic}  & TEXT-CNN & openSMILE        & 3D-CNN    \\
				MCTN \cite{pham2019found}    & GLOVE           & Facet        & COVAREP   & SACL-LSTM \cite{hu2023supervised}  & RoBERTa  & openSMILE        & 3D-CNN    \\
				MulT \cite{tsai2019multimodal}    & GLOVE           & Facet        & COVAREP   & HyCon  \cite{mai2022hybrid}     & BERT     & Facet            & COVAREP   \\
				MAG \cite{rahman2020integrating}     & BERT            & Facet        & COVAREP   & HGraph-CL \cite{lin2022modeling}   & BERT     & Facet            & COVAREP   \\
				ICDN \cite{zhang2023icdn}     & GLOVE           & Facet        & COVAREP   & bc-LSTM \cite{poria2017context}     & TEXT-CNN & 3D-CNN           & openSMILE \\
				AMOA \cite{li2022amoa}    & BERT            & OpenFace 2.0 & openSMILE & MMMU-BA  \cite{ghosal2018contextual}   & GLOVE    & Facet            & COVAREP   \\
				ICCN \cite{sun2020learning}    & BERT            & Facet        & COVAREP   & MISA  \cite{hazarika2020misa}       & BERT     & Facet            & COVAREP   \\ \bottomrule
	\end{tabular}}}
\end{table*}

\subsubsection{Text Feature Extraction}
With the rapid development of deep learning, word embedding has also been widely used to extract text features. {Before embedding, raw text data typically undergoes several preprocessing steps to improve the quality and consistency of the input. These steps often include tokenization, lowercasing, punctuation and stop-word removal, and in some cases, lemmatization or stemming. Some studies also perform syntactic or dependency parsing to capture the structural relationships between words, which can further enhance the semantic representation in downstream tasks.} Word embedding technology then uses a shallow neural network to learn the semantic information of words, and uses Euclidean distance to measure the similarity between words. {Unlike traditional one-hot encoding methods, word embedding technology maps high-dimensional sparse feature vectors to low-dimensional dense vectors. This reduces computational resource usage and addresses the issue of one-hot encoding's inability to capture the semantic gap between words.} A commonly used word embedding method is {word to vector (Word2Vec)} \cite{chen2019audio}, which contains two different forms: {continuous bag of words (CBOW)} \cite{ghosh2021depression} and Skip-gram \cite{du2020cross}. CBOW predicts the central word based on the surrounding words, and Skip-gram predicts the surrounding words based on the central word. Although the above methods can capture the semantic similarity between words, they require large datasets for training.

Some recent studies use TextCNN \cite{kim2014convolutional} and {global vectors for word representation (GLOVE)} \cite{gan2022semglove} to extract text features. In addition, large-scale predictive pre-training models such as {bidirectional encoder representations from Transformers (BERT)} \cite{ma2021t} and {robustly optimized BERT approach (RoBERTa)} \cite{kim2021randomly} are often used to capture contextual information through attention mechanisms.

\subsubsection{Video Feature Extraction}
Visual feature extraction is mainly to extract information such as facial expressions and gestures that contain the speaker's emotions from the video. In recent years, deep neural networks have been able to extract deep features from images in an end-to-end learning manner, avoiding the tedious manual feature extraction. For example, Tran et al. \cite{tran2015learning} proposed an effective and efficient 3D-CNN to process video frames containing spatio-temporal features.

{In most modern multimodal emotion recognition systems, visual preprocessing begins with frame sampling, where video is typically downsampled to a fixed frame rate (e.g., 25 or 30 frames per second) to reduce redundancy and maintain temporal resolution \cite{kattenborn2021review, wang2020cnn}. Each frame is then resized (commonly to 224×224 pixels) and normalized using per-channel mean subtraction and standard deviation scaling (e.g., ImageNet normalization settings) to ensure consistency across inputs. For facial region detection and alignment, face detectors (e.g., dlib or MTCNN) are employed to locate the face in each frame, and affine transformations are applied to align facial landmarks to a canonical pose, improving robustness to head movements and variations in scale or rotation. Various open-source toolkits are used to extract deep visual features from the aligned facial regions. For instance, OpenFace 2.0 \cite{baltrusaitis2018openface} detects 68 facial landmarks, estimates head pose, facial action units (AUs), gaze direction, and eye-blink frequency, which are strongly correlated with affective states. Facet further extracts features such as facial muscle activation, histograms of oriented gradients (HOG), emotion intensity scores, and micro-expressions. These features are typically calculated on a frame-by-frame basis and then aggregated over time using statistical functions (e.g., mean, standard deviation, max) or temporal models such as LSTMs. OKAO Vision is a commercial toolkit capable of estimating smile intensity (ranging from 0 to 100) and eye gaze orientation, while CERT adaptively captures head pose dynamics and subtle expressions over short temporal segments. These visual feature extractors are configured with default or task-optimized parameters depending on the specific dataset conditions (e.g., lighting, camera angle, facial occlusion).}

\subsubsection{Audio Feature Extraction}
Deep learning has increasingly attracted attention in the field of audio feature extraction, enabling automatic modeling of acoustic patterns associated with human emotion. For instance, LSTM networks \cite{xie2019speech} have been widely applied to model temporal dependencies in speech, while Poria et al. \cite{poria2017context} used convolutional neural networks (CNN) to extract local patterns from audio signals, followed by feeding the extracted features into emotion classification models.

In recent years, an increasing number of emotion recognition models \cite{tzinis2018integrating, majumder2019dialoguernn, zhong2019knowledge} have adopted open-source toolkits for systematic and standardized audio feature extraction. Commonly used tools include COVAREP \cite{dumpala2023manifestation}, openSMILE \cite{kumar2022fake}, LibROSA \cite{suman2022visualization}, and OpenEAR \cite{schepker2020acoustic}. These toolkits provide frame-level acoustic descriptors based on well-established speech analysis techniques. {Specifically, OpenEAR is capable of extracting a comprehensive set of low-level descriptors, such as prosodic (e.g., pitch, energy), spectral, and cepstral features, and applies Z-score normalization to ensure feature comparability across samples. The openSMILE toolkit is often configured with the INTERSPEECH 2010 or eGeMAPS feature set, extracting Mel-Frequency Cepstral Coefficients (MFCC), pitch, zero-crossing rate, voice intensity, and other prosody-based features. Audio signals are typically resampled to 16 kHz mono channel, then segmented using sliding windows (e.g., 25 ms with 10 ms stride) to generate frame-wise features. LibROSA, a widely used Python-based audio analysis library, is used to extract 33 frame-level acoustic features, including 20-dimensional MFCCs, chroma features, and Constant-Q Transform (CQT), which captures tonal energy variations. Similarly, COVAREP provides features such as 12-dimensional MFCC, Maxima Dispersion Quotient (MDQ), Normalized Amplitude Quotient (NAQ), and Liljencrants–Fant (LF) glottal model parameters, which are valuable for capturing subtle vocal tract changes related to emotional state.}


\section{Taxonomy of Multi-modal Conversational Emotion Recognition Algorithms}
\label{sec:sec4}
In this section, we present a taxonomy of MCER modeling approaches. We categorize existing work into context-free modeling, sequential context modeling, distingguishing speaker modeling, and speaker relation modeling. We briefly introduce each method in the following.

\subsection{Context-free Modeling}
These are mostly pioneering works on conversational emotion recognition. Context-free modeling methods aim to learn a feature representation for each sentence, which does not exploit the contextual information of the sentence {\cite{zhang2020emotion, seng2016combined, lotfian2019curriculum}}. For example, some traditional machine methods (e.g., SVMs \cite{rozgic2012ensemble, lin2005speech}, and decision trees \cite{cichosz2007emotion, lee2011emotion}, etc) is used to extracts the feature representation of each sentence, and utilize the extracted sentence features to complete emotion classification. The above process assumes that each sentence is independent and does not influence each other. We introduce several common context-free modeling methods based on feature fusion below.

\subsubsection{Add}
The early fusion method based on addition operation obtains the final emotional feature representation by weighted summation of different modality features \cite{deng2021survey}. This fusion method is simple to operate and requires only a small amount of calculation. However, its shortcomings are also obvious. It cannot model the context information in a fine-grained manner, and the information that can be utilized is limited. The formula for implementing the context-free modeling method using the additive approach is defined as follows:
\begin{equation}
	h_e = x^t + x^a + a^v
\end{equation}
where $h_e$ represents the fused emotional vectors, $x^t, x^a, x^v$ represent the text, audio, and video vectors, respectively. {The Add method is essentially the direct accumulation of different modal information in the same semantic space, which retains the weighted contribution of each modality in the corresponding feature dimension. Through summation, the model can automatically adjust the numerical expression of each modal feature during the learning process, making the information complementary and improving the overall representation ability.}

{The Add method is relatively simple and easy to understand and implement. By directly merging the features of different modalities, the feature information of different modalities can be fully utilized. For modalities with strong complementarity, the Add method can well capture the correlation between them. However, the features of different modalities may have different scales or importances. The Add method does not consider the difference in importance between the features of different modalities, which may lead to information loss or imbalance. Therefore, the Add method can be considered in scenarios where there is strong complementarity between modalities, limited computing resources, or requirements for model complexity.}

\subsubsection{Concatenation}
The early fusion method based on concatenation operation obtains the final emotion feature representation by concatenating and merging different modal features \cite{cambria2018benchmarking}. Although this fusion method does not introduce additional calculations, it leads to very high dimensionality of the data, which makes calculations difficult. Furthermore, it also fails to capture intra-modal and inter-modal semantic information that is complementary.
\begin{equation}
	h_e = Concat\left([x^t, x^a, a^v]\right)
\end{equation}
where $Concat\left( \cdot \right)$ represents concatenation operation. {The concatenation method does not perform any information compression or mapping on the features of each modality, but directly concatenate the original information by dimension, theoretically retaining the complete expression of each modality.}

{The concatenate method only requires simple vector concatenation operations, without complex parameter learning or model design. The concatenate method avoids possible information loss in early fusion. However, the feature dimension after concatenation is the sum of each modality, which may lead to sparsity and overfitting, especially on small-scale data sets. In addition, high-dimensional features may increase the computational complexity of subsequent models and simple concatenation cannot explicitly model the nonlinear relationship between modalities. Therefore, when the information quality of each modality is high and complementary to each other, concatenation can effectively retain information.}

\subsubsection{SVM}
SVM is a machine learning algorithm for classification and regression whose optimization goal is to find a hyperplane (a straight line in two-dimensional space, and a hyperplane in high-dimensional space) that separates samples of different classes. Based on the above research, Perez-Rosas et al. \cite{perez2013utterance} concatenate multi-modal features as input vectors and use SVM to classify utterances for emotion. SVM works better for binary classification problems, but is less effective in multi-classification problems, and is only suitable for training small-scale data sets. The formula of SVM is defined as follows:
\begin{equation}
	f\left(x\right)=sign\left(\sum_{i=1}^{N}\alpha_{i}^{*}y_{i}\exp\left(-\frac{\|x-z\|^{2}}{2\sigma^{2}}\right)+b^{*}\right)
\end{equation}
where $sign(x > 0) = 1, sign(x = 0) = 0, sign(x < 0) = 1$, $\alpha_{i}^{*}, b^{*}$ represents the learnable parameters, $\exp\left(-\frac{\|x-z\|^{2}}{2\sigma^{2}}\right)$ represents the kernel function, $N$ is the number of the samples. {SVM uses a nonlinear kernel function to implicitly model the nonlinear interaction information between modalities.}

{Through the kernel function, SVM can find the best classification hyperplane in the high-dimensional feature space, effectively perform nonlinear classification, and is suitable for high-dimensional data. When there are fewer training samples, SVM has good generalization ability, but it is time-consuming for large-scale data sets.}

\subsubsection{Multiple Kernel Learning}
After preprocessing the features of three different modalities, Poria et al. \cite{poria2015deep} constructed two different feature selectors to achieve feature dimensionality reduction. One of the feature selectors is based on circular correlated feature subset selection (CFS), and the other is based on principal component analysis (PCA). The above two feature selectors can not only eliminate redundant information and noise information, but also improve the running speed of the model. After feature selection and dimensionality reduction, the researchers spliced and merged the processed feature vectors and trained a classifier using a multi-kernel learning (MKL) algorithm \cite{poria2015deep}. Based on the previous research work, the authors further propose the convolutional recurrent multi-kernel learning (CRMKL) \cite{poria2016convolutional} model. CRNKL uses a convolutional recurrent neural network for emotion detection, which can extract contextual information. The formula of MKL is defined as:
\begin{equation}
	\begin{gathered}\max_{\alpha,\beta}\left[\sum_{i=1}^N\alpha_i-\sum_{i,j=1}^N\alpha_i\alpha_jy_iy_j\mathrm{K}_{mkl}(x_i,x_j)\right]\\ \sum_i^N\alpha_iy_i=0\\0\leq\alpha_i\leq C\\
		\mathrm{K}_{mkl}=\sum_{k}^{M}\beta_kK_k > 0
	\end{gathered}
\end{equation}
where $y_i$ is the true label, $\alpha, \beta$ are the learnable parameters, $M$ is the feature dimension. {MKL achieves flexible fusion of multimodal information at the kernel space level through multi-core combination and weight optimization.}

{The multiple kernel learning method can combine multiple different kernel functions according to different data characteristics, which helps to process complex data structures. It is suitable for processing a variety of heterogeneous data or multimodal data and can combine information from different modalities. However, calculating the combination of multiple kernel functions may result in high computational costs.}

\begin{figure*}
	\centering
	\includegraphics[width=1\linewidth]{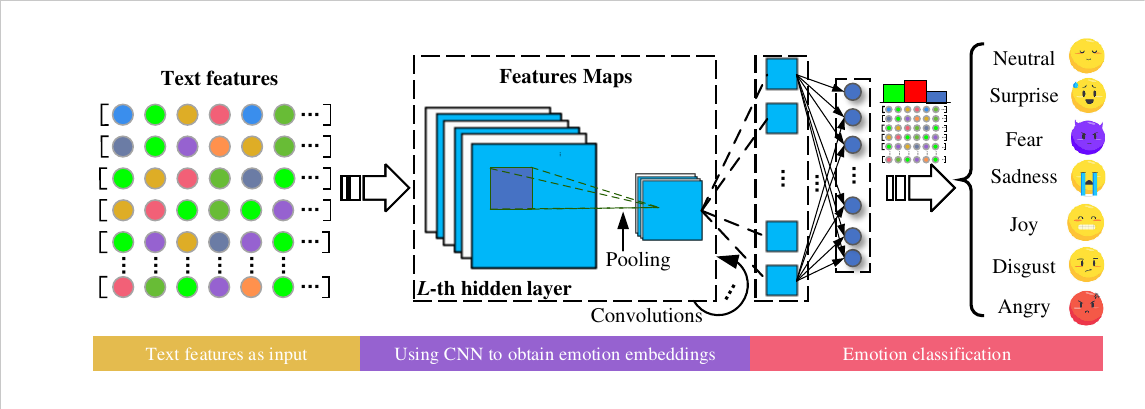}
	\caption{The flowchart of the proposed TextCNN method. Specifically, given text features, the TextCNN employs convolution filters of different sizes to generate feature maps, and uses 1D-max pooling to expand the receptive field of feature maps, and further utilizes a multi-layer perceptron (MLP) to complete emotional prediction.}
	\label{fig:cnn}
\end{figure*}

\subsubsection{Select-Additive Learning CNN}
CNN is a classic neural network in visual tasks and cannot be directly used for emotion recognition. To solve this problem, Kim et al. \cite{kim2014convolutional} proposed the TextCNN model, and its overall process is shown in Fig. \ref{fig:cnn}. To perform multi-modal emotion recognition, Wang et al. \cite{wang2017select} proposed the SAL-CNN model, which first uses multi-modal data to fully train the CNN, and then uses Select-Additive Learning (SAL) to improve its versatility and prevent the model from overfitting during training. The SAL method consists of two phases (i.e., selection and addition). In the selection phase, SAL preserves important features and removes noisy information from the latent feature representations learned from neurons. In the addition phase, SAL improves the model's noise immunity by adding Gaussian noise to the feature representation. The SAL method improves the generalization performance of deep fusion models.

The formula for extracting text features by CNN is defined as follows:
\begin{equation}
	\begin{aligned}
		x_{1:n}^t = x_1 \oplus x_2 \oplus \ldots x_n \\
		c_i = f(\omega \cdot x_{p:p+q-1} + b)
	\end{aligned}
\end{equation}
where $\oplus$ represents concatenation operator, $\omega$ represents convolution filter, $c_i$ represents the feature representation within a window, $f(\cdot)$ represents activation function. Convolutional filters are used to extract features from all sentences to generate feature maps:
\begin{equation}
	\mathbf{c}=maxpooling[c_1,c_2,\ldots,c_{n-h+1}]
\end{equation}
The max pooling operation is used to capture the most critical semantic information in the sentence.

It can be seen from the processing flow of the convolutional neural network that using CNN to extract text features does not contain contextual information, i.e., it is assumed that each sentence is independent of each other.

The CNN model is relatively simple and has a fast training speed. It can effectively extract local features in text, especially when processing long texts. Although CNN handles long texts well, it has poor adaptability to short texts and structured data and may lose some word order information when processing texts with sequential order.

\subsubsection{Tensor Fusion Network}

\begin{wrapfigure}{r}{0.43\textwidth} 
	\centering 
	\includegraphics[width=0.88\linewidth]{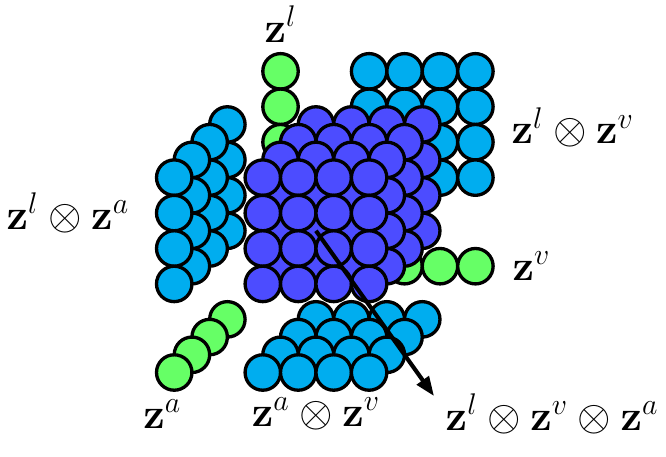}
	\caption{{Illustration of the Tensor Fusion Network (TFN) for tri-modal fusion. The feature vectors from the language ($\mathbf{z}^l$), acoustic ($\mathbf{z}^a$), and visual ($\mathbf{z}^v$) modalities are first augmented with a constant term and then combined via tensor outer product. This operation explicitly captures unimodal, bimodal (e.g., $\mathbf{z}^l \otimes \mathbf{z}^a$, $\mathbf{z}^a \otimes \mathbf{z}^v$), and trimodal ($\mathbf{z}^l \otimes \mathbf{z}^v \otimes \mathbf{z}^a$) interactions in a structured tensor space.}}
	\label{fig:3models-mosipaper} 
\end{wrapfigure}

As shown in Fig. \ref{fig:3models-mosipaper}, the tensor-based feature fusion method mainly calculates the tensor product of different modal feature representations through Cartesian product to obtain the fused tensor representation \cite{pandey2022attention}. Therefore, the above methods need to first map the input multi-modal feature representation into a high-dimensional space, and then map it back to a low-dimensional tensor space for emotion representation. Tensor-based methods are able to capture important high-order interaction information across time, space, and modality. However, the computational complexity of tensor methods is very high and grows exponentially, and there is no fine-grained semantic information interaction between modalities. Zadeh et al. \cite{zadeh2017tensor} proposed the multi-modal tensor fusion network (TFN). TFN adopts the method of tensor fusion, which can simulate the interaction process between the three modalities of text, audio and video, and effectively fuse multi-modal features. Although TFN can effectively model information interaction within and between modalities, the model complexity of the TFN method is related to the dimensionality of multi-modal features and grows exponentially. The formula of TFN is defined as follows:
\begin{equation}
	\left.\left\{(x^t,x^v,x^a)\mid x^t\in\begin{bmatrix}\mathbf{x}^l\\1\end{bmatrix}\right.,x^v\in\begin{bmatrix}\mathbf{x}^v\\1\end{bmatrix},x^a\in\begin{bmatrix}\mathbf{x}^a\\1\end{bmatrix}\right\}
\end{equation}
where the extra dimension with 1 is used to perform modal interaction. The Cartesian product is then used to fuse the three modal features as follows:
\begin{equation}
	\begin{aligned}
		\mathbf{x}^m & = \mathbf{x}^l \otimes \mathbf{x}^v \otimes \mathbf{x}^a \\
		& =\begin{bmatrix}
			1 & \mathbf{z}_a^\top \\
			\mathbf{z}_v & \mathbf{z}_v \mathbf{z}_a^\top
		\end{bmatrix}
		\begin{bmatrix}
			\mathbf{z}_l & \mathbf{z}_l \mathbf{z}_a^\top \\
			\mathbf{z}_l \mathbf{z}_v^\top & \mathbf{z}_l \mathbf{z}_v^\top \mathbf{z}_a^\top
		\end{bmatrix}
	\end{aligned}
\end{equation}
where $\otimes$ represents the outer product, $x^m$ represents fused vectors. {With the help of tensor outer products, the interaction information of all levels can be systematically preserved.}

{Tensor fusion methods use tensor decomposition and high-dimensional fusion techniques to map multimodal information into a unified space and capture high-order relationships between modalities through tensor operations. However, the computational overhead of processing tensor operations is high, which may lead to computational bottlenecks during training.}

\subsubsection{Low-rank Tensor Fusion Network}
\begin{figure}
	\centering
	\includegraphics[width=1\linewidth]{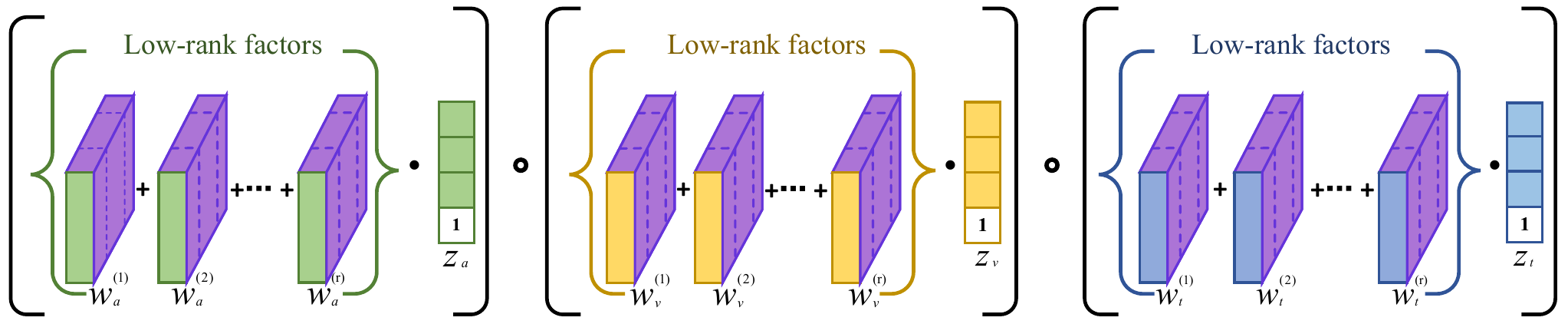}
	\caption{The overall flow chart of LFM. LFM mainly performs low-rank decomposition of the learnable parameters of specific factors of the mode.}
	\label{fig:low-rank}
\end{figure}

On the basis of TFN, in order to more efficiently fuse multi-modal data, Liu et al. \cite{liu2018efficient} proposed a low-rank tensor fusion (LFM) method to achieve dimensionality reduction of multi-modal features, so as to improve the fusion efficiency of multi-modal features as shown in Fig. \ref{fig:low-rank}. LFM has achieved high performance on many different tasks.
\begin{equation}
	\begin{aligned}
		\mathbf{x}^m& =\left(\sum_{i=1}^r\mathbf{w}_a^{(i)}\otimes\mathbf{w}_v^{(i)} \otimes\mathbf{w}_t^{(i)}\right)\cdot\mathbf{x}  \\
		&=\left(\sum_{i=1}^r\mathbf{w}_a^{(i)}\cdot x_a\right)\circ\left(\sum_{i=1}^r\mathbf{w}_v^{(i)}\cdot x_v\right)\circ\left(\sum_{i=1}^r\mathbf{w}_t^{(i)}\cdot x_t\right)
	\end{aligned}
\end{equation}
where $\mathbf{w}_a, \mathbf{w}_v, \mathbf{w}_t$ represents the decomposed low-rank learnable tensor. {LFM essentially retains the multimodal high-order information expression capability of tensor outer product and explicitly models the joint distribution and deep semantic dependencies of different modalities through low-rank approximation.}

{Low-rank Tensor Fusion Network reduces the complexity of the model by low-rank decomposition of tensors, which can significantly reduce the computational overhead of tensors and reduce memory requirements. However, although low-rank decomposition can reduce model complexity, it may lose some information.}

\subsubsection{Data Augmentation with  Generative Adversarial Networks}
Multimodal emotion recognition based on adversarial learning is an advanced direction in this field, which combines the principles of adversarial learning to improve the accuracy and robustness of emotion recognition {\cite{ren2023maln, yuan2023noise}}. Next, we introduce the existing overall process of data augmentation based on adversarial generative networks.

%

\textbf{1) Conditional GANs}
Conditional Generative Adversarial Network (cGAN) \cite{sun2023discriminatively} is a variant of GAN that introduces conditional information to more precisely control the output of the generator. The core idea of cGAN is to pass additional condition information to the generator and discriminator during the generation process, thereby generating specific types of data based on given conditions. The main advantage of cGAN is its ability to precisely control the generation process in order to generate data that meets the conditional information. The optimization goal of cGAN is defined as:
\begin{equation}
	\begin{gathered}
		\min_G\max_DV(D,G) 
		=\mathbb{E}_{\mathbf{x}\sim p_{\mathrm{data}}(\mathbf{x})}\{\log D([\mathbf{x},\mathbf{y}])\} 
		+\mathbb{E}_{\mathbf{z}\sim p_z(\mathbf{z})}\{\log(1-D([G([\mathbf{z},\mathbf{y}]),\mathbf{y}]))\} 
	\end{gathered}
\end{equation}
where $x$ represents real data, and $y$ represents extra information. 

\begin{equation}
	\begin{aligned}
		&\mathcal{L}_{D}^{(cGAN)} =-\mathbb{E}_{\mathbf{x}\sim p_{\mathrm{data}}(\mathbf{x})}\{\log D([\mathbf{x},\mathbf{y}])\}  -\mathbb{E}_{\mathbf{z}\sim p_{z}(\mathbf{z})}\{\log(1-D([G([\mathbf{z},\mathbf{y}]),\mathbf{y}]))\} \\
		&\mathcal{L}_{G}^{(cGAN)} =-\mathbb{E}_{\mathbf{z}\sim p_z(\mathbf{z})}\{\log(D([G([\mathbf{z},\mathbf{y}]),\mathbf{y}]))\}
	\end{aligned}
\end{equation}

\textbf{2) Adversarial Autoencoders}
Adversarial Autoencoder (AAE) \cite{latif2020multi} combines the ideas of autoencoder and GAN. The main goal of AAE is to make this encoding space more continuous and have better data generation capabilities while learning a compressed representation of data. The training objective function of AAE usually includes two parts: one is the reconstruction error of the autoencoder, which ensures the quality of the encoding, and the other is the GAN loss, which makes the encoding distribution more continuous and closer to the real distribution. The formula is defined as follows:
\begin{equation}
	\begin{aligned}
		&\mathcal{L}_{D}^{(AAE)} =-\mathbb{E}_{\mathbf{z}\sim p_z(\mathbf{z})}\{\log D(\mathbf{z})\}  -\mathbb{E}_{\mathbf{x}\sim p_{\mathrm{data}}(\mathbf{x})}\{\log(1-D(E(\mathbf{x})))\} \\
		&\mathcal{L}_{E}^{(AAE)} =-\mathbb{E}_{\mathbf{x}\sim p_{\mathrm{data}}(\mathbf{x})}\{\log(D(E(\mathbf{x})))\}  \\
		&\mathcal{L}_{R}^{(AAE)} =\mathbb{E}_{\mathbf{x}\sim p_{\mathrm{data}}(\mathbf{x})}\big\{||\mathbf{x}-R(E(\mathbf{x}))||^{2}\big\} 
	\end{aligned}
\end{equation}
where $p_z(z)$ represents the prior distribution. 

\textbf{3) Adversarial Data Augmentation Network}
Adversarial data augmentation network (ADAN) \cite{wang2022m2r2} includes the following components: autoencoder $R(E(x))$, auxiliary classifier $C(E(x))$, generator $G(z, y)$ and discriminator $D(h)$. First, ADAN aims to learn a latent representation of the input data $x$ to preserve the emotional information in it. Second, it attempts to ensure that the generated latent representation is consistent with the emotional information of the input data by matching the posterior distribution $p(h|z, y)$ with $p(h|x)$. Third, ADAN simultaneously strives to minimize the reconstruction error between the input data $x$ and its reconstructed version $\hat{x}$ to ensure high-quality data reconstruction. The generator $G(z, y)$ accepts a sample $z$ drawn from an M-dimensional Gaussian distribution and a one-hot encoding of the emotion label $y$ as input, and the goal is to generate samples in the latent space such that they are indistinguishable from real samples. The discriminator $D(h)$ is optimized to distinguish whether the latent vector $h$ comes from real data or from the generator.
\begin{equation}\begin{aligned}
		&\mathcal{L}_{D}^{(\mathrm{ADAN})} =-\mathbb{E}_{\mathbf{x}\sim p_{\mathrm{data}}(\mathbf{x})}\{\operatorname{log}D(E(\mathbf{x}))\}  -\mathbb{E}_{\mathbf{z}\sim p_z(\mathbf{z})}\{\log(1-D(G(\mathbf{z},\mathbf{y})))\} \\
		&\mathcal{L}_{C}^{(\text{ADAN})} =-\mathbb{E}_{\mathbf{x}\sim p_{\mathrm{data}}(\mathbf{x})}\bigg\{\sum_{k=1}^{K}y_{\mathrm{emo}}^{(k)}\log C(E(\mathbf{x}))_k\bigg\}  \\
		&\mathcal{L}_{R}^{(\mathrm{ADAN})} =\mathbb{E}_{\mathbf{x}\sim p_{\mathrm{data}}(\mathbf{x})}\{||\mathbf{x}-R(E(\mathbf{x}))||^2\}  \\
		&\mathcal{L}_{E}^{(\mathrm{ADAN})} =\mathbb{E}_{\mathbf{x}\sim p_{\mathrm{data}}(\mathbf{x})}\left\{||\mathbf{x}-R(E(\mathbf{x}))||^2 -\sum_{k=1}^Ky_{\mathrm{emo}}^{(k)}\log C(E(\mathbf{x}))_k\right\} \\
		&\mathcal{L}_G^{\mathrm{(ADAN)}}=\mathbb{E}_{\mathbf{z}\sim p_z(\mathbf{z})}\bigg\{\log(1-D(G(\mathbf{z},\mathbf{y})))-\alpha\sum_{k=1}^Ky_{\mathrm{emo}}^{(k)}\log C(G(\mathbf{z},\mathbf{y}))_k\bigg\}
\end{aligned}\end{equation}
where $\alpha$ determines the contribution of classification error to model optimization. {ADAN improves the authenticity of fused features and the ability to align cross-modal distributions through the confrontation between the discriminator and the generator.}

{Generative adversarial networks are able to generate high-quality new samples that are very close to real data through an adversarial training process. This enables the model to generate new samples with greater diversity and authenticity, thereby improving the model's generalization ability. For tasks with scarce data, especially when there are fewer samples of a specific category, GANs can be used to enhance the dataset by generating new samples, avoiding the limitations of traditional data augmentation methods on data diversity and complexity. However, the training process of GANs is often unstable, and the adversarial process between the generator and the discriminator may cause the gradient to vanish or explode, resulting in unstable quality of the generated samples.}

\subsection{Sequential Context Modeling}
\begin{figure*}
	\centering
	\includegraphics[width=1\linewidth]{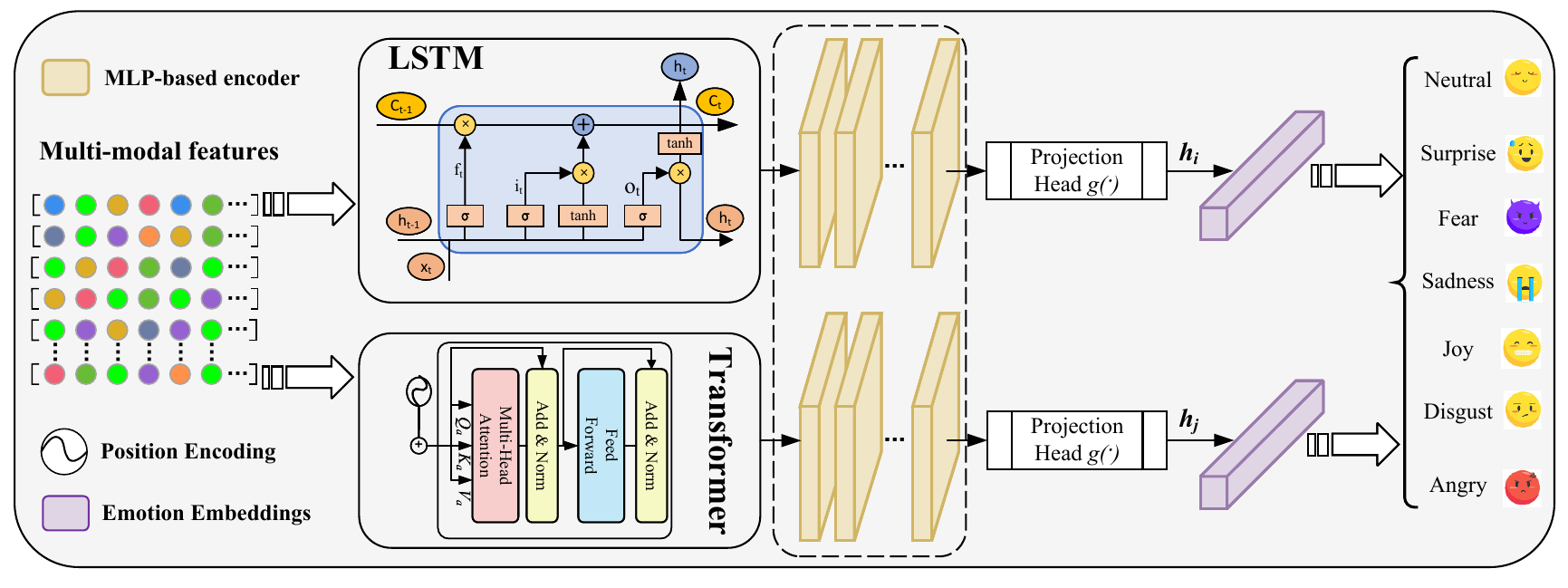}
	\caption{The flowchart of the proposed contextual modeling approach. Sequential context modeling methods use LSTMs or Transformer to capture high-level features with rich contextual semantic information among different modal features.}
	\label{fig:contextual-modeling}
\end{figure*}

Context-free modeling is conceptually important and has inspired later research on sequential context modeling \cite{tu2022context}. In particular, sequential context modeling methods consider that contextual sentences are mutually influential. Sequential context modeling approaches \cite{ma2019emotion, tao2018advanced, xie2019speech} consider each sentence influenced by its surrounding utterances. The main idea is to generate a feature representation with rich contextual semantic information by combining its own utterance representation $x_i$ with the surrounding contextual sentence representation $\{x_{i-k}, \cdots, x_{i-1}, x_{i+1}, \cdots, x_{i+k}\}$, where $k$ represents the context window size. Different from the context-free modeling method, the sequential context modeling method obtains a better feature representation by setting a memory network to preserve the context information of the sentence. Taking Fig. \ref{fig:contextual-modeling} as an example, a LSTM or Transformer is used to extract contextual information for three modalities of video, audio and text. The sequential context modeling approach plays an important role for many other MCER modeling approaches.

{Tri-modal Hidden Markov Model is a sequence context modeling method, which relies on the previous state and can effectively capture local dependencies.} For example, Morency et al. used text, video, and audio features for the task of trimodal emotion analysis, and designed a model to extract useful information in different modal features \cite{morency2011towards}. After extracting multi-modal features, the three modal features are connected and input into a Hidden Markov Chain (HMM) classifier \cite{morency2011towards} to learn the emotional state of the input signal. {HMM believes that the state of the current moment is only related to the information of the previous moment, which enables the model to use the context information of the utterance.} The formula of HMM is defined as follows:
\begin{equation}
	\begin{aligned}
		P(w|x^a,x^v, x^t)& =\sum_{i=1}^C\sum_{j=1}^D\sum_{k=1}^M P(w,\lambda_i^a,\lambda_j^v, \lambda_k^t|x^a,x^v, x^t) \\
		&=\sum_{i=1}^C\sum_{j=1}^D\sum_{k=1}^MP(w|\lambda_i^a,\lambda_j^v, \lambda_k^t, x^a,x^v, x^t) \\
		&\times P(\lambda_i^a,\lambda_j^v, \lambda_k^t|x^a,x^v, x^t)
	\end{aligned}
\end{equation}
where $C, D, M$ represent feature vector dimensions for audio, video, and text, $w$ represents the emotional class, $P(\lambda_i^a,\lambda_j^v, \lambda_k^t|x^a,x^v, x^t)$ represents the confidence of the emotion classification.

Since the true class label is based on the output of the predicted class label $\hat{w}_b$, the formula of HMM can be expanded as follows:
\begin{equation}
	\begin{aligned}
		&	P(w|\lambda^a_i,\lambda^v_j, \lambda^t_k, x^a,x^v, x^t) \\ & =\sum_{b=1}^B P(w,\hat{w}_b|\lambda_i^a,\lambda_j^v, \lambda_k^t, x^a,x^v, x^t) \\
		&=\sum_{i=1}^BP(w|\hat{w}_b, \lambda_i^a,\lambda_j^v, \lambda_k^t, x^a,x^v, x^t) \\
		&\times P(\hat{w}_b|\lambda_i^a,\lambda_j^v, \lambda_k^t,x^a,x^v, x^t)
	\end{aligned}
\end{equation}
where $P(\hat{w}_b|\lambda_i^a,\lambda_j^v, \lambda_k^t,x^a,x^v, x^t)$ represents the probability of predicted label. {HMM captures cross-modal temporal dependencies and synchronization relationships through temporal dependency structures.}

{HMM can effectively process sequence data, especially when there are unobservable hidden states at each time step. By modeling the hidden state, HMM can capture the time dependency in the data. Therefore, HMM has strong capabilities in sequence modeling. However, HMM is based on linear models and cannot effectively handle complex nonlinear relationships in sequence data. For nonlinear sequence data, the performance of HMM may be limited. In addition, when processing large-scale data, the computational complexity of HMM training is high, especially when the parameter space is large, computational bottlenecks may be encountered.}

LSTM is a variant of RNN that can remember contextual information. Specifically, LSTM models long-distance dependent context through cellular units and can solve the vanishing gradient problem. Each LSTM consists of input gate $j_t$, output gate $O_t$, cell state ${C}_t$, and forget gate $f_t$.
\begin{equation}\begin{aligned}
		&\begin{bmatrix}\widetilde{C}_t\\O_t\\j_t\\f_t\end{bmatrix}=\begin{bmatrix}\tanh\\\sigma\\\sigma\\\sigma\end{bmatrix}W_T\begin{bmatrix}x_t\\h_i^{t-1}\end{bmatrix} \\
		&C_t=C_t\odot j_t+C_{t-1}\odot f_t \\
		&h_{i}^{t}=O_{t}\odot\tanh(C_{t})
\end{aligned}\end{equation}
where $\sigma$ represents activation function. {LSTM extracts dynamic information within the modality by capturing time dependencies.}

{LSTM can remember information over long time spans and is particularly suitable for modeling long-term dependencies. However, LSTM still performs worse than other more advanced models (such as Transformer) in some extremely long sequence tasks. LSTM's ability to retain memory over long time spans is also limited. Therefore, LSTM is suitable for most tasks that need to capture time dependencies, but for very long dependencies or large-scale data, modern models such as Transformer may provide better performance.}

After LSTM was used in multi-modal conversational emotion recognition, many other works were proposed to extract contextual emotional information. Lu et al. \cite{lu2023exploring} proposed a multi-scale LSTM multi-modal emotion recognition model, which uses LSTM to extract low-level and high-level local emotional features in multi-modal features. This method can capture subtle changes in complex expressions in a more fine-grained manner and implement an information feedback mechanism. However, it cannot capture the status information of the utterance and the status information of the speaker.

Existing models ignore modal alignment and directly fuse information on different modal features. Modal alignment can eliminate the heterogeneity of single-modal features and obtain accurate emotional representations of different modal features. Based on this current situation, Hou et al. \cite{10050091} proposed a semantic alignment network based on multi-space learning, which uses LSTM to extract emotional features of different modalities and obtains high-level emotional representations as supervisory signals for modal alignment. This method can capture the global correlation between different modalities and achieve feature fusion between modalities.

Transformers are another way of modeling sequential context \cite{li2020hitrans, zhu2021topic, huang2020multimodal}. Transformer's long-distance modeling capabilities are far superior to recurrent neural networks, and Transformer can achieve parallel computing. Therefore, existing research on multi-modal emotion recognition based on sequential context modeling often regards Transformer as an important technology. The implementation details of Transformer are as follows.

Firstly, video, audio and text features (i..e., $x^t, x^a, x^v$) are concatenated into a fusion vector. The formula is defined as follows:
\begin{equation}
	Q, K, V = Concat(x^t, x^a, x^v)
\end{equation}
where $Q, K, V$ represent the query vector, key vector and value vector of multi-modal features, respectively. {Transformer automatically captures key modal and key area information by dynamically assigning weights.}

Secondly, we use a feedforward neural network to perform multiple linear transformations on $Q$, $K$, and $V$. The formula is defined as follows:
\begin{equation}
	\begin{aligned}
		& \tilde{Q} =Concat(QW_{1}^{Q},\ldots,QW_{i}^{Q},\ldots,QW_{m}^{Q}) \\
		& \tilde{K} =Concat(KW_{1}^{K},\ldots,KW_{i}^{K},\ldots,KW_{m}^{K}) \\
		& \tilde{V} =Concat(VW_{1}^{V},\ldots,VW_{i}^{V},\ldots,VW_{m}^{V})
	\end{aligned}
\end{equation}
where $m$ represents the number of linear transformations.

We then perform multi-head attention in parallel to obtain emotion feature representation:
\begin{equation}
	\begin{aligned}
		&head_{i} =\frac{softmax\left((QW_{i}^{Q})(KW_{i}^{K})^{T}\right)}{VW_{i}^{V}}  \\
		&H_{head} =Concat(head_1,\ldots,head_m)
	\end{aligned}
\end{equation}
where $H_{head}$ represents the emotion feature vectors.

Finally, position encoding is used to obtain the position information of the emotion sequence:
\begin{equation}
	\begin{aligned}
		PE_{(pos,2i)}&=\sin\left(\frac{pos}{10000^{2i/d}}\right)\\PE_{(pos,2i+1)}&=\cos\left(\frac{pos}{10000^{2i/d}}\right)
	\end{aligned}
\end{equation}
where pos is the index of the $i$-th sentence, position encoding information is fused into $Q$, $K$, and $V$.

{Transformer uses a self-attention mechanism to process input data, which is independent of the order of the sequence. Compared with traditional RNNs and LSTMs, Transformer can calculate the relationship between each input position in parallel, which significantly speeds up the training process. Due to its parallelism, Transformer is more suitable for training large-scale datasets and can greatly reduce training time. In addition, Transformer's Multi-head Self-Attention allows the model to focus on different parts of the input from multiple perspectives, which helps capture multi-level information. However, since Transformer involves interactions with all positions when calculating self-attention, its time complexity and space complexity are both high, especially when processing long sequences, the amount of calculation increases quadratically. For long sequences, Transformer's computational and memory requirements are very large. In addition, when the amount of data is limited, Transformer may not be able to fully exert its advantages, requiring a large amount of pre-training data and careful tuning.}

After Transformer was proposed, many Transformer-based multi-modal conversational emotion recognition methods were proposed to model long-distance context dependencies \cite{gerczuk2021emonet, hazarika2021conversational}. {Previous works failed to model long-distance dependencies between different modal features. To address this, Yang et al. \cite{9905904} proposed a multimodal speech emotion recognition method using a context Transformer, which enhances the emotional representation of the current utterance by embedding contextual information.} This method can adaptively learn feature fusion between modalities.

{Existing methods struggle to dynamically identify subtle emotional changes in multimodal and multi-scale features. To address this, Liu et al. \cite{liu2022multi} proposed a multi-scale self-attention fusion emotion recognition method, which uses the self-attention mechanism to extract context-related dependencies in multimodal features.} Therefore, there is potential to use Transformers to model long-distance context dependencies. This method combines bc-LSTM and a multi-head attention mechanism to achieve fine-grained emotional information mining, and uses feature-level fusion and decision-level fusion methods to experiment with cross-modal feature fusion.

\begin{figure*}
	\centering
	\includegraphics[width=1\linewidth]{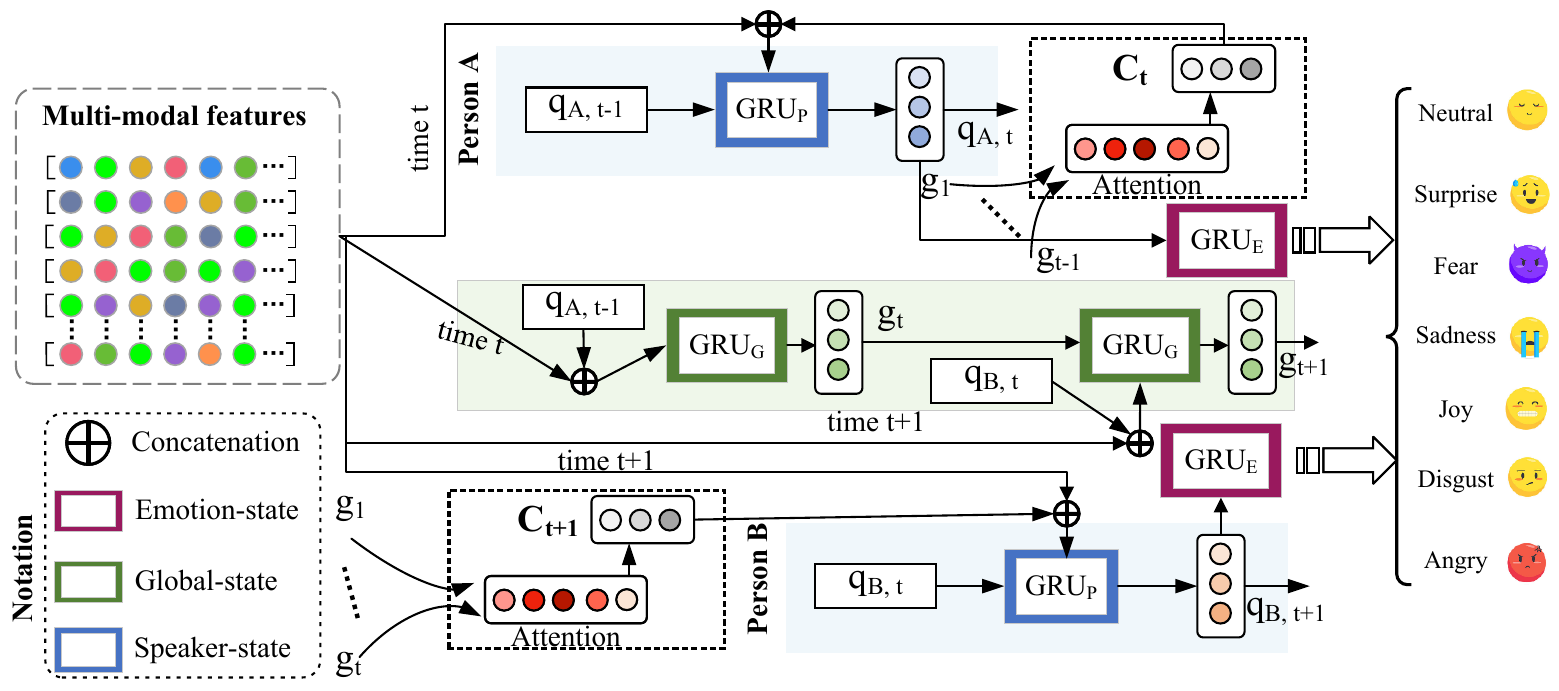}
	\caption{The flowchart of the proposed Distingguishing speakers modeling approach. The distinguishing speakers modeling approach designs three GRU states, i.e., a global GRU, an emotional GRU and a speaker GRU, which are used to update global context information, emotion category information and speaker information, respectively.}
	\label{fig:distinginsting}
\end{figure*}

\subsection{Distinguishing speaker modeling}
The distinguishing speakers modeling method considers that the speaker's emotion is not only related to the global context, but also related to the speaker's own emotional state. Take Fig. \ref{fig:distinginsting} as an example, there are three GRU states (i.e., a global GRU, an emotional GRU and a speaker GRU). The global GRU is utilized to extract global multi-modal information and speaker's emotional state information. The speaker GRU is used to fuse the semantic information with context captured by the attention mechanism and the speaker's emotional state information. The emotion GRU combines the speaker's emotional state information and global context information to complete the final emotion classification.

Global GRU captures the contextual semantic information of an utterance by modeling the utterance and speaker states. Each speaker state is used to memorize a speaker-specific representation of an utterance. By distinguishing the subordination relationship between speakers and utterances, it is beneficial to model the dependency relationship between speakers and utterances, thereby enhancing the semantic representation ability of context. The formula of Global GRU is defined as:
\begin{equation}
	g_t=GRU_{\mathcal{G}}(x_{t-1},(x_t\oplus q_{s(x_t),t-1}))
\end{equation}
where $g_t$ represents the latent feature representation of the global state, $q_{s(x_t)}$ represents the speaker state of the current utterance $x_t$.

Speakers typically reply to conversations based on contextual information from other. Therefore, speaker GRU extracts the context $c_t$ related to the utterance $x_t$. The formula is defined as follows:
\begin{equation}
	\begin{aligned}
		&\beta  =\text{softmax}(x_t^TW_\beta[g_1,g_2,\ldots,g_{t-1}]),  \\
		&c_{t} =\beta[g_{1},g_{2},\ldots,g_{t-1}]^{T}
	\end{aligned}
\end{equation}
where $W_\beta$ is the learnable parameters. First, calculate the attention score of the global state in the previous $t-1$ time. The attention score assigns higher weight to utterances related to utterance $x_t$. The final context vector $c_t$ is obtained by the dot product of the attention score $\beta$ and the global state $g_t$.

\begin{equation}
	q_{s(u_t),t}=GRU_{\mathcal{P}}(q_{s(u_t),t-1},(u_t\oplus c_t))
\end{equation}

The emotional representation et of the utterance $u_t$ is obtained by combining the speaker's state $q_{s(ut),t}$ and the utterance $e_{t-1}$ at time $t-1$. One underlying intuition is that context has a greater impact on utterance $u_t$, and $e_{t-1}$ integrates emotional contextual information from other parties’ states into the emotional representation $e_t$. Therefore, we use the Emotion GRU unit to model $e_{t_1}$, and the formula is defined as follows:
\begin{equation}
	e_t=GRU_{E}(e_{t-1},q_{s(u_t),t})
\end{equation}
The emotion representation $e_t$ that combines context information and speaker status information is used for the final emotion classification. {Distinguishing speaker modeling method combines global context with individual information to improve understanding of complex information such as semantics and emotions.}

{Similar to traditional RNN and LSTM models, distinguishing speakers modeling methods have the ability to capture long-range dependencies. In multi-turn conversations, the previous conversation content often affects subsequent understanding and generation. Distinguishing speakers modeling methods can effectively learn these long-term dependencies and help understand contextual information. However, in scenarios where the amount of data is small or the conversation data is simple, distinguishing speakers modeling methods may suffer from overfitting. In particular, when there are not many conversation rounds in the training data set, the model tends to remember specific patterns in the training set and cannot effectively generalize to new data.}

{In modeling methods based on distinguishing between speakers, Ghosal et al. \cite{ghosal2020cosmic} proposed {commonsense knowledge for emotion identification in conversations (COSMIC)}, which clarifies the relationship between the speaker and the utterance. It also introduces common sense knowledge to enhance the emotional understanding of the model.} COSMIC can learn a variety of different prior knowledge (e.g., event relationships and causal relationships, etc.), and can distinguish speaker information and dynamically detect the speaker's emotional changes.

In view of the fact that existing methods cannot pay attention to the correlation between utterances and speakers and the lack of interaction between speakers, Zhang et al. \cite{zhang2021coin} proposed a conversational interaction model, which extracts contextual semantic information and state interaction information of utterances through stacked global interaction modules. In addition, this method also implements adversarial feature representation of the model by introducing noise information. Experimental results prove that adversarial learning can improve the performance of emotion recognition.

\begin{figure*}
	\centering
	\includegraphics[width=1\linewidth]{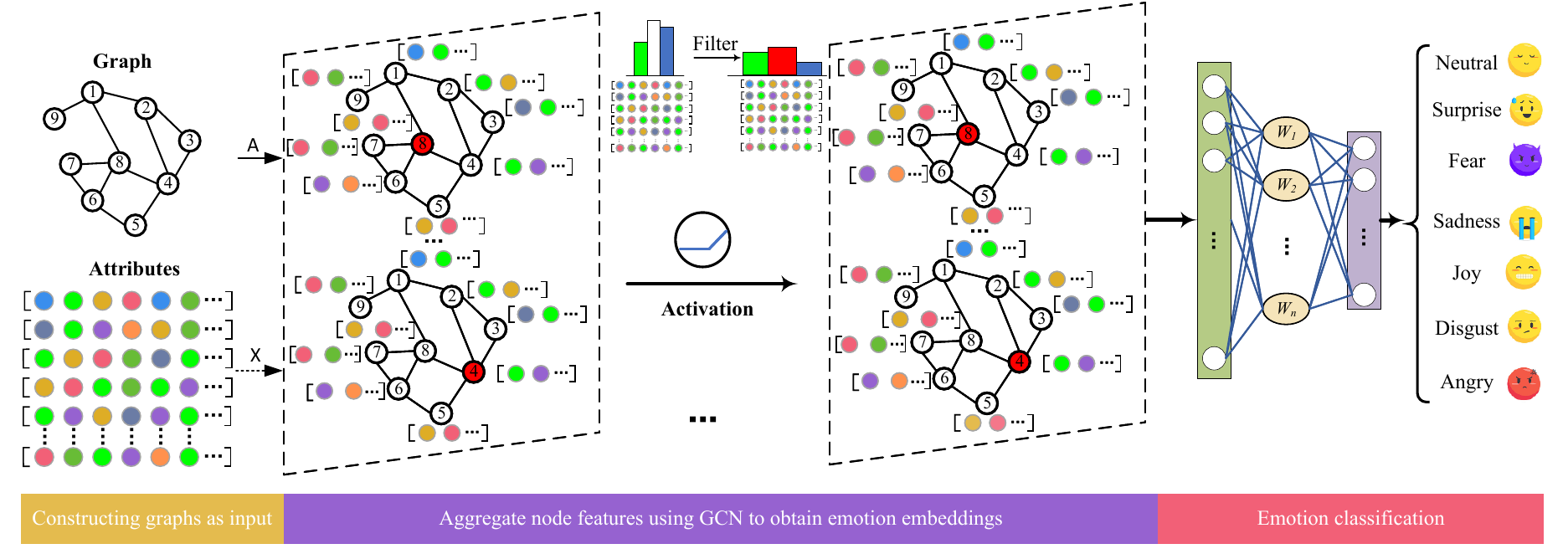}
	\caption{The flowchart of the proposed speaker relationship modeling approach. The speaker modeling method aggregates dialogue relationship information between speakers by using a graph convolutional neural network.}
	\label{fig:speaker-modeling}
\end{figure*}

\subsection{Speaker Relationship Modeling}
\subsubsection{GNN for speaker relationship modeling}
The speaker relationship modeling method innovatively introduces graph neural network to capture the speaker's dialogue relationship information while extracting sequential context information. Taking Fig. \ref{fig:speaker-modeling} as an example, it extracts dialogue relationships between speakers and inter-speaker dependencies by constructing a speaker relationship graph.

{Graph convolutional network (GCN)} extends convolution operations into graph-structured data to extract structural information. GCN performs first-order neighbor information aggregation and spectral domain estimation. The formula of GCN is defined as follows:
\begin{equation}
	\label{eq:gcn}
	\boldsymbol{H}^{(l+1)}_i=ReLU\left(\tilde{D}^{-\frac12}\tilde{A}\tilde{D}^{-\frac12}\boldsymbol{H}^{(l)}\boldsymbol{W}^{(l)}\right)
\end{equation}
where $\boldsymbol{W}^{(l)}$ is the learnable parameters, $\tilde{A}=A +I_n$, $I_n$ is the identity matrix, $\tilde{\boldsymbol{D}}_{ii}=\sum_j\tilde{a}_{ij}$. $\boldsymbol{H}^{(l+1)}$ represents the latent feature representations of layer $l+1$.

The steps to apply GCN to the field of multi-modal emotion recognition are as follows. First, each utterance is represented as a node in the graph, and edge relationships are constructed based on the context between utterances. We then apply GCN to the constructed dialogue graph for speaker-level information extraction. Through the above process, the model can dynamically learn the correlation between sentences. According to the definition of Equation \ref{eq:gcn}, our formula for aggregating surrounding contextual utterence information is deformed as follows:
\begin{equation}
	\begin{aligned}
		H_{i}^{(l+1)} =ReLU\left(\sum_{r\in\mathcal{R}}\sum_{j\in\mathcal{N}_i^r}\frac1{|\mathcal{N}_i^r|}\left(W_{\theta_1}^{(l)}H_j^{(l)} +W_{\theta_2}^{(l)}H_i^{(l)}\right)\right)
	\end{aligned}
\end{equation}
where $W_{\theta_1}$ and $W_{\theta_2}$ are the learnable parameters, $\mathcal{N}_i^r$ represents the neighbor node under the relationship $r \in \mathcal{R}$.

{The multi-layer convolutional structure of GCN can effectively integrate feature information from different modalities and enhance the expressiveness of emotional information. Especially when the amount of information is large, GCN can effectively aggregate the features of each modality, which helps the accuracy of emotional classification or regression tasks. However, GCN has difficulties in modeling long-distance dependencies. As the depth of the graph increases, the problem of over-smoothing will occur during the information propagation process, causing the representation of nodes to become similar and lose their original distinguishability.}

{Graph Attention Network (GAT)} is a variant of GCN that aggregates surrounding neighbor node features through learnable weights with an attention mechanism. GAT captures the more important node features in the graph by calculating the degree of similarity between nodes. The formula for GAT is defined as follows:
\begin{equation}
	\boldsymbol{H}_i^{(l+1)}=ReLU\left(\sum_{j\in N(w_i)}\alpha_{ij}^{(l+1)}\boldsymbol{W}^{(l+1)}\boldsymbol{h}_j^{(l)}\right)
\end{equation}
where $\alpha_{ij}$ is the edge weight between node $i$ and node $j$.

Similarly, the formula for using GAT to extract conversational relationships between speakers is defined as follows:
\begin{equation}
	\begin{aligned}
		H_{i}^{+(l+1)} =ReLU\left(\sum_{r\in\mathcal{R}}\sum_{j\in\mathcal{N}_i^r}\frac1{|\mathcal{N}_i^r|}\left(\alpha_{ij}^{(l)}W_{\theta_1}^{(l)}H_j^{+(l)}\right.\right.  +\left.\alpha_{ii}^{(l)}W_{\theta_2}^{(l)}H_i^{+(l)}\right)
	\end{aligned}
\end{equation}
{GAT dynamically distinguishes the importance of neighboring nodes, giving the model stronger nonlinear expression and local pattern capture capabilities.}

{In multimodal emotion recognition, the relationship between modalities may be heterogeneous, that is, the data of different modalities may be different in nature. GAT can effectively handle such heterogeneous graphs because it gives different weights to each type of node through the attention mechanism, thereby better representing the heterogeneous relationship between different modalities. However, although the attention mechanism can improve the flexibility of information aggregation, it may also lead to excessive focus on local information, especially in some tasks that need to consider the global context. The local weighting of GAT may limit the global learning ability of the model.}

The multi-modal method based on GNN is the current mainstream research, which can consider context information and speaker relationship information simultaneously \cite{ghosal2019dialoguegcn}. {To jointly learn sequential context, multimodal interaction, and multitask representation, Zhang et al. \cite{zhang2023m3gat} designed the M3GAT (Multi-modal, Multi-task Interactive Graph Attention Network). M3GAT simultaneously models context dependencies, multimodal emotional interactions, and speaker dependencies. It enables cross-modal feature interaction, captures sequential contextual semantic information, and establishes task correlations.}

\begin{figure}
	\centering
	\includegraphics[width=0.98\linewidth]{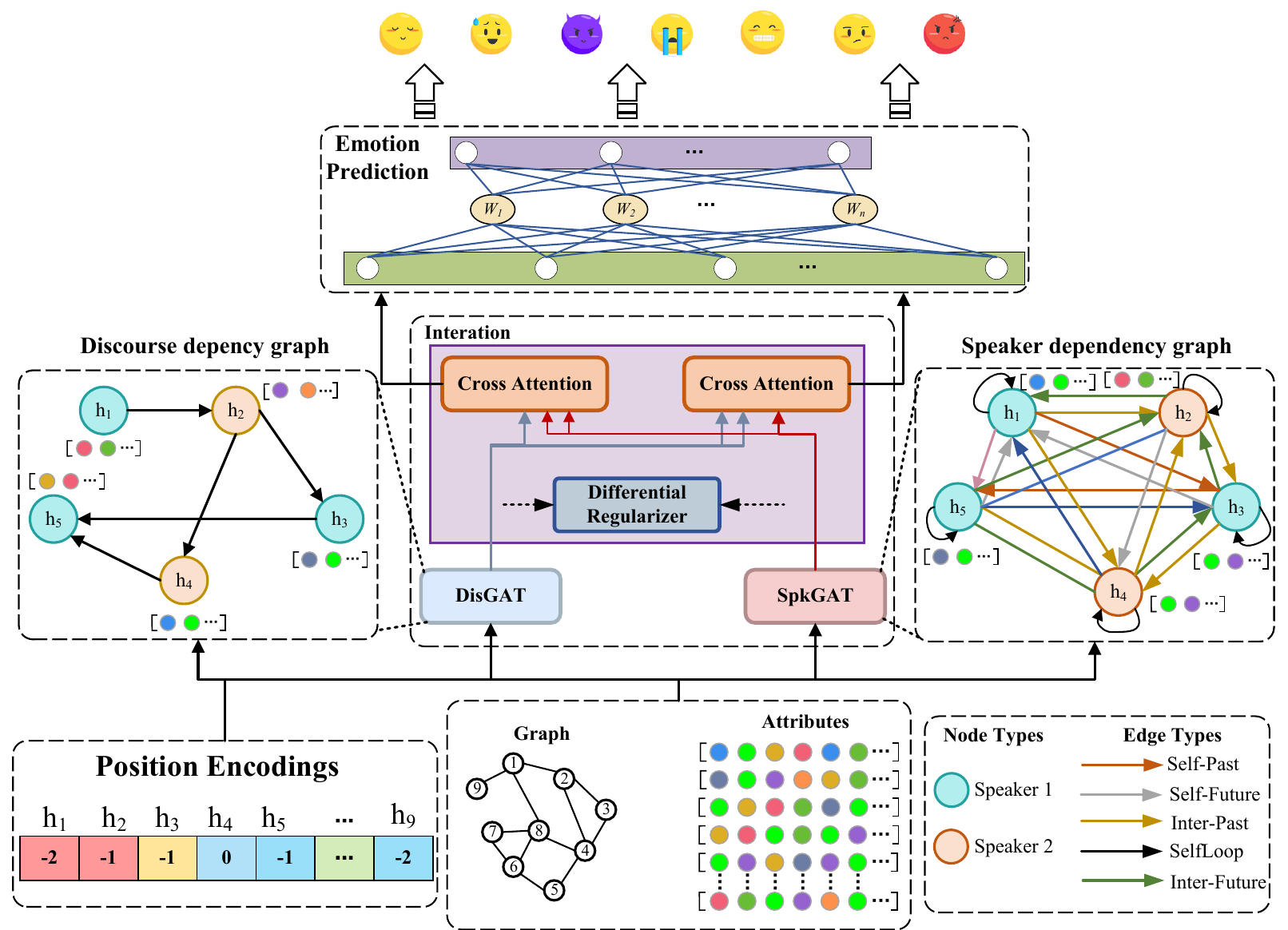}
	\caption{The flow chart of RGAT. RGAT mainly includes dialogue relationship dependency graph, speaker dependence graph and position coding information.}
	\label{fig:rgat}
\end{figure}

{Existing graph fusion methods often cause the model to lose important semantic information and fail to eliminate redundancy. To address this, Li et al. \cite{li2023graphcfc} proposed a graph network based on cross-modal feature complementarity.} This method effectively extracts the speaker's context and interaction information using multiple hypothesis spaces in the graph. This method eliminates the heterogeneity between modalities and fuses modal information by performing different message aggregation on different nodes and edge relationships in the graph, thereby extracting contextual information and speaker relationship information.

Although existing MCER methods use GCN to model conversational relationships between speakers. In particular, the most competitive methods model the dependence of conversational relations between speakers and the importance between conversational relations by using relational graph attention networks. However, existing GCN-based multimodal conversational emotion recognition methods do not consider conversational relationships and sequential information in contextual relationships. Based on the above problems, Ishiwatari et al. \cite{ishiwatari2020relation} introduced relational position coding in {relationship graph attention network (RGAT)} to provide sequence information. The specific flow chart of RGAT is shown in Fig. \ref{fig:rgat}.

The position encoding formula used by RGAT is defined as follows:
\begin{equation}\left.PE_{ijr}=\left\{\begin{array}{rl}max(-p,min(p,j-i))&r=1,wherej\in\mathcal{N}^1(i)\\max(-p,min(p,j-i))&r=2,wherej\in\mathcal{N}^2(i)\\max(-f,min(f,j-i))&r=3,wherej\in\mathcal{N}^3(i)\\max(-f,min(f,j-i))&r=4,wherej\in\mathcal{N}^4(i)\end{array}\right.\right.\end{equation}
where $PE_{ijr}$ represents the relative position distance between node $i$ under relationship type $r$ and its surrounding neighbor nodes $j$. The maximum relative position distance between nodes is clipped to $p$ or $4$ , which represents the context window size. $\mathcal{N}r(i)$ represents the neighborhood of node $i$ under relationship type $r$. To make the position encoding information learnable, feedforward network (FFN) is used to obtain position embeddings.

{RGAT improves emotion recognition performance by explicitly modeling complex relationships within and between modalities. However, if the prior relationship definition is unreasonable (e.g., incorrectly connecting irrelevant modalities), the performance may deteriorate.}

\subsection{Emotion Classification}
After obtaining the multi-modal emotion feature representation, the MCER task uses a multi-layer perceptron and a softmax layer to achieve the final emotion classification. The probability distribution of emotion categories is as follows:
\begin{equation}
	\begin{aligned}
		&l_{t} =\operatorname{ReLU}(W_{l}e_{t}+b_{l})  \\
		&\mathcal{P}_{t} =\text{softmax}(Wl_{t}+b)  \\
		&\hat{y}_{t} =\underset{i}{\operatorname*{argmax}}(\mathcal{P}_{t}[i])
	\end{aligned}
\end{equation}
where $W_l, W, b_l, b$ are the learnable parameters, $\mathcal{P}_{t}$ is the probability distribution of emotion categories, $\hat{y}_{t}$ is the predicted labels.

\section{Evaluation Metrics}
\label{sec:sec5}
For MCER tasks, there are four commonly used evaluation indicators, i.e., accuracy rate, weighted average accuracy rate (WA), F1 value, and weighted average F1 value (WF1). These four indicators are defined as follows:

We assume that $N$ is the number of emotion labels in the dialogue emotion dataset, $E_j$ represents the total number of samples of emotion labels in the $j$-th, $j \in [1,N]$.

1) Accuracy represents the emotion recognition accuracy of the model, and the formula is defined as follows:
\begin{equation}
	\operatorname{Accuracy}_{j}=\frac{\sum_{n=1}^{\vartheta_{2}} E_{j}^{i}}{\sum_{m=1}^{\vartheta_{1}} S_{j}^{m}}
\end{equation}
where $\vartheta_1$ is the number of labels on a certain category of emotion. $\vartheta_2$ is the number that the model predicts on a certain category of emotion. $E_j^i$ means that the $i$-th sample in the $j$-th emotionally predicted correctly. $E_j^i \in [0,1]$. $S_j^m$ represents the $m$-th sample of the $j$-th emotion. The larger the value of $Accuracy_j$, the better the recognition effect of the model on the $j$-th type of emotion.

2) The F1 value is the F1-score of each emotion, and the formula is defined as follows:
\begin{footnotesize}
	\begin{equation}
		F 1_{j}=\frac{2 \times \operatorname{Recall}\left(E_{T P}^{j}, E_{F P}^{j}\right) \times \operatorname{Precision}\left(E_{T P}^{j}, E_{F N}^{j}\right)}{\operatorname{Recall}\left(E_{T P}^{j}, E_{F P}^{j}\right)+\operatorname{Precision}\left(E_{T P}^{j}, E_{F N}^{j}\right)}
	\end{equation}
\end{footnotesize}
and
\begin{equation}
	\begin{gathered}
		\operatorname{Precision}\left(E_{T P}^{j}, E_{F N}^{j}\right)=\frac{\left|E_{T P}^{j}\right|}{\left|E_{T P}^{j} \cup E_{F N}^{j}\right|} \\
		\operatorname{Recall}\left(E_{T P}^{j}, E_{F P}^{j}\right)=\frac{\left|E_{T P}^{j}\right|}{\left|E_{T P}^{j} \cup E_{F P}^{j}\right|}
	\end{gathered}
\end{equation}
where $E_{TP}^j$ is the number of samples that the model predicts correctly on the $j$-th category of emotion, $E_{FP}^j$ is the number of samples that the model predicts incorrectly on the $j$-th category of emotion, and $E_{FP}^j$ is the number of emotions from other categories that the model predicts as the $j$-th category of emotion. $Precision(E_{TP}^j,E_{FN}^j)$ is the model's precision on the $j$-th category of emotion, and $Recall(E_{TP}^j,E_{FP}^j)$ is the recall of the model on the $j$-th emotion. f1 value combines the effects of both precision and recall metrics. Usually, the larger the value of f1, the better the prediction of the model.

3) Weight accuracy (WA) is the weighted average of the classification accuracy of all emotion categories. The more samples of the $j$-th emotion, the smaller the weight of the sample. The formula is defined as follows:
\begin{equation}
	W A=\frac{\sum_{m=1}^{\vartheta_{1}} S_{j} * \text { Accuracy }_{j}}{\sum_{j=1}^{N} \sum_{m=1}^{\vartheta_{1}} S_{j}^{m}}
\end{equation}
WA is the classification accuracy of the model combining all emotions. The larger the WA, the better the model performs on average across all classes.

4) Weight F1 (WF1) is the weighted F1 value of all emotion categories. The more samples of the $j$-th emotion, the smaller the weight of the sample. The formula is defined as follows:
\begin{equation}
	W F 1=\frac{\sum_{m=1}^{\vartheta_{1}} S_{j} * F 1_{j}}{\sum_{j=1}^{N} \sum_{m=1}^{\vartheta_{1}} S_{j}^{m}}
\end{equation}

WF1 is the F1 value where the model integrates all emotions. WF1 is another effective index to evaluate the model effect. In general, the larger the WF1, the better the average performance of the model across all classes.

\begin{table*}[htbp]
	\centering
	\caption{We count the performance of different types of emotion recognition algorithms on publicly available datasets. The weighted F1 score is chosen as evaluation metric.}
	\label{table:algorithms}
	\renewcommand\arraystretch{1}
	\scalebox{0.7}{
		\setlength{\tabcolsep}{3.5mm}{
			\begin{tabular}{lcccc}
				\toprule
				Approaches       & Category                                  & Inputs & Database                          & Performence(\%)     \\ \midrule
				SAL  \cite{wang2017select}        & Context free                     & T+A+V      & IEMOCAP/MELD                      & 49.2/58.8           \\
				SVM \cite{rozgic2012ensemble}             & Context free                    & T+A+V      & IEMOCAP/MELD                      & 48.7/56.4           \\
				TFN \cite{zadeh2017tensor} & Context free & T+A+V      & IEMOCAP/MELD &  54.2/56.7  \\
				LFM \cite{liu2018efficient} & Context free & T+A+V      & IEMOCAP/MELD &  55.3/56.7  \\
				UniMSE \cite{hu2022unimse}          & Sequential context                & T+V+A  & IEMOCAP/MELD                      & 70.7/65.5           \\
				bc-LSTM+Att \cite{poria2017context}    & Sequential context              & T+V+A  & IEMOCAP/MELD                      & 55.0/56.4           \\
				M2FNet  \cite{chudasama2022m2fnet}         & Sequential context                & T+V+A  & IEMOCAP/MELD                      & 69.9/66.7           \\
				CESTa \cite{wang2020contextualized}            & Sequential context               & T+V+A  & IEMOCAP/DailyDialog/MELD          & 67.1/63.1/58.4      \\
				CMN  \cite{hazarika2018conversational}            & Sequential context               & T+V+A  & IEMOCAP                           & 56.2                \\
				SACL-LSTM \cite{hu2023supervised}       & Sequential context                &    T+A+V    & IEMOCAP/MELD/EmoryNLP             & 69.2/66.5/39.7      \\
				Att-BiLSTM \cite{tzinis2018integrating} & Sequential context               & T+V+A  & IEMOCAP                           & 62.9                \\
				DialogueCRN \cite{hu2021dialoguecrn}      & Sequential context               & T+A+V      & IEMOCAP/MELD                      & 66.2/58.39          \\
				EmoCaps \cite{li2022emocaps}         & Sequential context                & T+V+A  & IEMOCAP/MELD                      & 71.8/64.0           \\
				ICON \cite{hazarika2018icon}             & Sequential context               & T+V+A  & IEMOCAP                           & 63.5                \\
				DialogueRNN \cite{majumder2019dialoguernn}      & Distinguishing speakers  & T+V+A  & IEMOCAP/MELD                      & 62.8/56.8           \\
				EmotionIC \cite{yingjian2023emotionic}        & Distinguishing speakers  & T+V+A  & IEMOCAP/DailyDialog/MELD/EmoryNLP & 69.5/59.8/66.4/40.0 \\
				COIN \cite{zhang2021coin}             & Distinguishing speakers  & T+V+A  & IEMOCAP                           & 65.4                \\
				COSMIC  \cite{ghosal2020cosmic}          & Distinguishing speakers  & T+A+V      & IEMOCAP/DailyDialog/MELD/EmoryNLP & 65.3/58.5/65.2/38.1 \\
				RGAT \cite{ishiwatari2020relation}             & Speaker relationship              & T+A+V      & IEMOCAP/DailyDialog/MELD/EmoryNLP & 65.2/54.3/60.9/34.4 \\
				DialogueGCN \cite{ghosal2019dialoguegcn}     & Speaker relationship              & T+V+A  & IEMOCAP/MELD                      & 64.2/58.1           \\
				DAG-ERC \cite{shen2021directed}         & Speaker relationship              & T+A+V      & IEMOCAP/DailyDialog/MELD/EmoryNLP & 68.0/59.3/63.7/39.0 \\
				MM-DFN  \cite{hu2022mm}         & Speaker relationship              & T+V+A  & IEMOCAP/MELD                      & 68.2/59.5           \\
				GraphCFC \cite{li2023graphcfc}        & Speaker relationship              & T+V+A  & IEMOCAP/MELD                      & 68.9/58.9           \\
				
				\bottomrule
	\end{tabular}}}
\end{table*}

\begin{table*}[htbp]
	\centering
	\caption{{We count the performance of different types of emotion recognition algorithms on publicly available datasets. The precision, recall, and AUC are chosen as evaluation metric.}}
	\label{table:metri}
	\renewcommand\arraystretch{1}
	\scalebox{0.7}{
		\setlength{\tabcolsep}{3mm}{
			\begin{tabular}{lcccc}
				\toprule
				Approaches    & Database                                  & Precision(\%) & Recall(\%)                          & AUC(\%)     \\ \midrule
				SAL  \cite{wang2017select}  &  IEMOCAP/MELD         &  51.3/60.2     &  50.3/57.4                    &   70.9/76.6        \\
				SVM \cite{rozgic2012ensemble}   &  IEMOCAP/MELD            &   49.1/59.4    &  47.6/57.9                  &   65.3/73.2     \\
				TFN \cite{zadeh2017tensor}    & IEMOCAP/MELD      & 57.3/57.0  &   55.4/56.6 & 76.1/72.9     \\
				LFM \cite{liu2018efficient} &   IEMOCAP/MELD      & 56.2/58.4    &  57.3/56.1    & 78.4/71.7     \\
				UniMSE \cite{hu2022unimse}  &   IEMOCAP/MELD      &  68.8/65.2            & 65.4/64.7 &    83.6/79.0                         \\
				bc-LSTM+Att \cite{poria2017context}  &   IEMOCAP/MELD    &    56.7/58.9           &  57.4/56.3 &   76.4/71.7                            \\
				M2FNet  \cite{chudasama2022m2fnet} &  IEMOCAP/MELD       &   67.6/67.4            &  65.3/66.1 &    85.1/79.3                         \\
				CESTa \cite{wang2020contextualized}    & IEMOCAP/DailyDialog/MELD       &   68.5/64.3/59.2    & 67.2/65.7/62.3  &    87.9/76.7/79.2             \\
				CMN  \cite{hazarika2018conversational}  &   IEMOCAP       &    58.4           & 57.3  &      75.3                                 \\
				SACL-LSTM \cite{hu2023supervised}   &  IEMOCAP/MELD/EmoryNLP     &   69.1/65.1/43.8            & 67.3/64.7/45.6      &   88.4/79.0/60.1                \\
				Att-BiLSTM \cite{tzinis2018integrating} &   IEMOCAP    & 64.5        &  62.7 &     80.3                                \\
				DialogueCRN \cite{hu2021dialoguecrn}      &  IEMOCAP/MELD  &   67.4/61.4    &    65.3/62.3 &   84.0/79.3                         \\
				EmoCaps \cite{li2022emocaps}         &   IEMOCAP/MELD   &  70.1/64.5       &  71.2/62.7 &   86.8/78.5                            \\
				ICON \cite{hazarika2018icon}  &  IEMOCAP         &    64.3           &  62.0 &   83.3                                   \\
				DialogueRNN \cite{majumder2019dialoguernn}      &  IEMOCAP/MELD  & 65.5/58.3  &  63.7/59.6     &    80.9/73.6                       \\
				EmotionIC \cite{yingjian2023emotionic}  &  IEMOCAP/DailyDialog/MELD/EmoryNLP    & 70.0/62.5/65.6/43.1   & 68.6/63.3/64.3/46.8  &  87.2/72.1/75.2/55.3  \\
				COIN \cite{zhang2021coin}     &   IEMOCAP     & 67.8  & 66.5 &    85.0                                       \\
				COSMIC  \cite{ghosal2020cosmic}  &   IEMOCAP/DailyDialog/MELD/EmoryNLP     & 66.3/60.2/63.1/43.3  &  64.2/63.1/60.9/44.5   &  85.2/74.8/82.2/60.9  \\
				RGAT \cite{ishiwatari2020relation}   &  IEMOCAP/DailyDialog/MELD/EmoryNLP        &  67.8/55.2/61.3/36.3           & 65.7/56.3/62.4/39.9    &   84.2/66.9/69.7/52.3   \\
				DialogueGCN \cite{ghosal2019dialoguegcn}  &  IEMOCAP/MELD  &    65.3/57.1      &  54.2/58.5 &    83.9/69.2                         \\
				DAG-ERC \cite{shen2021directed}     & IEMOCAP/DailyDialog/MELD/EmoryNLP   &   69.5/60.7/64.2/41.3      &  68.4/61.3/62.2/43.8     &  87.9/72.3/66.1/62.0  \\
				MM-DFN  \cite{hu2022mm}         &  IEMOCAP/MELD   &  67.2/62.5        &   66.4/63.1 &     87.2/77.6                          \\
				GraphCFC \cite{li2023graphcfc}  &  IEMOCAP/MELD    &   69.3/60.4          &  68.6/58.0 &     89.1/81.5                              \\
				\bottomrule
	\end{tabular}}}
\end{table*}

\section{Experimental Results}
\label{sec:sec6}

{To comprehensively evaluate the performance of different multimodal emotion recognition (MCER) methods, this experiment systematically analyzes the existing representative methods from multiple dimensions. First, from the perspective of overall performance, we statistically analyze the comprehensive performance of each method under the weighted F1. This section focuses on comparing the differences in the effects of different categories of methods (e.g., context free, sequential context, distinguishing speakers, and speaker relationship modeling), reflecting the advantages of introducing context information and speaker dependence mechanisms in improving recognition performance. Secondly, from the perspective of fine-grained performance, the comparison results of each method in terms of precision, recall, and AUC are further listed, and the stability and generalization ability of the model under different evaluation indicators are comprehensively measured. Meanwhile, to deeply analyze the actual application efficiency of the model, the parameter scale, inference time, and classification accuracy and F1 score of each method for different refined emotion categories (e.g., happy, sad, angry, excited, fear, etc.) are statistically analyzed. This section reveals the trade-off between parameter complexity and performance of different methods, and highlights the differences in the ability to maintain high-precision classification under small parameters and low latency conditions.}

As shown in Table \ref{table:algorithms}, we present the emotion recognition effects of different algorithms on multiple data sets. In particular, each algorithm uses multi-modal data, and we distinguish different MCER algorithms according to our classification method. {Experimental results show that context-free based algorithms have the worst performance because they contain the least semantic information and cannot obtain good emotional feature representation. The multi-modal conversational emotion recognition algorithm based on sequential context has significant performance improvement compared to the context-free algorithm. The performance improvement may be attributed to the sequential context algorithm's ability to model the dependencies between contexts and its ability to utilize context information to improve the feature representation of emotions. The emotion recognition effects of modeling methods based on distinguishing speaker relationships and sequential context modeling methods are similar, and both are better than context-independent modeling methods. The performance improvement may be attributed to the ability of the distinguishing speakers modeling method to dynamically capture the speaker status information of the utterance and integrate it into the emotion representation information. The modeling method based on speaker relationship has the best performance and is currently the most popular modeling method. The modeling method based on speaker relationship mainly constructs the dialogue relationship between speakers through the inherent properties of the graph structure, and extracts the dialogue relationship representation between speakers through GCN. In addition, the speaker relationship modeling method can also consider the dependency information of the sequential context simultaneously.}

{To better understand the performance differences among various MCER algorithms, we provide a detailed analysis of each model’s architecture, focusing on how structural design, parameter complexity, and feature fusion strategy influence emotion recognition effectiveness. TextCNN and LFM are classic context-free baselines. TextCNN uses convolutional layers on word embeddings without any sequential modeling or multimodal interaction mechanisms. LFM extends this by incorporating limited modality fusion but still lacks temporal modeling. Their poor performance (e.g., $<$50\% F1 on MELD) confirms that models ignoring context and interaction structures struggle to capture emotional semantics. bc-LSTM and bc-LSTM+Att represent sequential context models based on Bi-LSTM architectures. They process utterances in temporal order, enabling the capture of inter-utterance dependencies. The addition of attention mechanisms improves performance by focusing on emotionally salient parts. This explains the consistent performance improvement over context-free models, especially on IEMOCAP. A-DMN leverages a dynamic memory network, which not only captures sequential information but also performs iterative attention-based reasoning. Its moderate parameter size (7.39M) and good F1 performance indicate its strength in temporal reasoning while maintaining efficiency. DialogueRNN, a foundational distinguishing speakers model, models each speaker’s emotional state over time using GRUs and a global attention mechanism. It explicitly distinguishes speaker roles, which improves emotion tracking in multi-speaker conversations. Its solid performance on both datasets confirms the benefit of dynamic speaker state modeling. DialogueGCN further advances this by applying graph convolution on utterance nodes, capturing both temporal and speaker-specific dependencies. It constructs directed conversation graphs and updates node representations via GCN layers. This architecture achieves strong overall F1, especially in emotions requiring long-term relational modeling (e.g., frustration and sadness). MM-DFN and M2ETNet are multimodal transformer-based models that focus on advanced modality fusion strategies. MM-DFN integrates modality-specific feature streams using deep fusion networks, while M2ETNet adds temporal and modality-level attention. These models excel at learning fine-grained multimodal interactions, resulting in top-tier performance across most emotion categories. Their performance, however, comes at the cost of larger parameter sizes and inference time. EmoCaps introduces a capsule network-based structure to model intra-modal hierarchies and inter-modal routing, capturing subtle semantic features. It shows strong performance in excited and angry categories, which benefit from complex vocal and textual cues. CT-Net integrates context and speaker-specific cues using temporal attention and cross-modal interactions. It performs stably across emotions with moderate complexity (8.49M). LR-GCN, the most recent and complex model, constructs a speaker relationship graph with relational GCN layers. It jointly models sequential context, speaker identity, and emotion transition. Despite its large parameter size (15.77M) and long runtime (up to 147s on MELD), it consistently achieves the best F1-scores across datasets and emotion categories. Its strength lies in combining graph reasoning, speaker modeling, and contextual information in a unified framework. DisGCN focuses on distinguishing speakers via graph structure, while SumAggGIN applies hierarchical aggregation over utterances. These methods show reasonable performance, but are often outperformed by speaker relationship models, suggesting that modeling explicit speaker interaction via GCN yields more robust features. Models such as ICON and CNN focus only on sequential context or modality fusion without graph reasoning. Their lower performance suggests that lack of explicit speaker modeling limits their ability to capture dialogue dynamic.}

{To provide a more comprehensive assessment of model performance, we incorporate multiple evaluation metrics including Precision, Recall, and AUC, in addition to the Weighted F1-score. As shown in Table~\ref{table:metri}, we observe that Graph-based models generally outperform traditional context-free and sequential models across all three metrics. For instance, GraphCFC achieves the highest AUC (89.18\%) and also maintains strong Precision (69.96\%) and Recall (60.48\%), indicating both robust classification ability and balanced detection across emotion categories. Among the sequential context models, architectures like SACL-LSTM and DialogueGCN show competitive performance, with SACL-LSTM achieving a Recall of 76.47\% and AUC of 88.47\%. This suggests that incorporating temporal dependencies and attention mechanisms enhances the model’s sensitivity to subtle emotional cues. Speaker modeling methods (e.g., EmotiCon, RGAT, DialogueRNN) also perform well. Notably, EmotiCon achieves a high Recall of 81.63\% but slightly lower Precision, suggesting that the model favors recall-oriented decisions, potentially useful in applications where missing emotional signals is more critical than occasional false alarms. Interestingly, simpler models such as CNN or UniMSE display relatively low AUC and Precision, indicating that they struggle to make accurate decisions across varied threshold settings, reinforcing the importance of context and multimodal integration in MCER tasks.}

\begin{table*}[!t]
	\renewcommand\arraystretch{1}
	\centering
	\caption{{On the IEMOCAP dataset, we counted the parameters, running time, and emotion recognition effects of different MCER algorithms on different emotion categories. The best result in each column is in bold.}}
	\label{tab:IEMOCAP}
	\scalebox{0.75}{
		\setlength{\tabcolsep}{4pt}
		\begin{tabular}{l|cccccccc}
			\hline
			\multirow{3}{*}{Methods} & \multicolumn{6}{c}{IEMOCAP}                                                              \\ \cline{2-9}
			& \multirow{2}{*}{{Parmas.}}  &  \multirow{2}{*}{{Running time}}  &  Happy      & Sad        & Neutral    & Angry      & Excited    & Frustrated  \\ \cline{4-9}
			&     &    & Acc.  F1   & Acc.  F1   & Acc.  F1   & Acc.  F1   & Acc.  F1   & Acc.  F1    \\ \hline
			TextCNN \cite{kim2014convolutional}    	& \textbf{{0.47M}}    &     \textbf{{0.96s}}                 & 27.73  29.81 & 57.14  53.83 & 34.36  40.13 & 61.12  52.47 & 46.11  50.09 & 62.94  55.78  \\
			bc-LSTM  \cite{poria2017context}     	& {1.28M}    &   {2.16s}            & 29.16  34.49 & 57.14  60.81 & 54.19  51.80 & 57.03  56.75 & 51.17  57.98 & 67.12  58.97  \\
			bc-LSTM+Att \cite{poria2017context} 	& {2.17M}    &  { 2.59s}       & 30.56 35.63 & 56.73 62.09 & 57.55 53.00 & {59.41} 59.24 & 52.84 58.85 & 65.88 {59.41}   \\
			CMN  \cite{hazarika2018conversational}    	& {3.85M}    & {4.14s}                   & 25.01  30.34 & 55.96  62.45 & 52.81  52.36 & 61.77  59.88 & 55.59  60.24 & {71.16}  60.67 \\
			LFM \cite{liu2018efficient}    	&  {6.24M}   &   {6.23s}          & 25.63  33.14 & 75.71  78.83 & 58.52  59.21 & 64.77  65.26 & {80.21}  71.85 & 61.14  58.97 \\
			A-DMN \cite{xing2020adapted}   	& {7.39M}    &    {6.69s}          & 43.15  50.64 & 69.47  76.88 & 63.05  62.92 & 63.53  56.56 & \textbf{88.34}  77.91 & 53.34  {55.72} \\
			DialogueRNN \cite{majumder2019dialoguernn}   	& {15.17M}    &   {20.05s}           & 25.63  33.11 & 75.14  78.85 & 58.56  59.24 & 64.76  65.23 & 80.27  71.85 & 61.16  58.97  \\
			DialogueGCN   \cite{ghosal2019dialoguegcn}   	&  {12.92M}   &   {14.18s}         & 40.63  42.71 & \textbf{89.14  84.45} & 61.97  63.54 & 67.51  64.14 & 65.46  63.08 & 64.13  66.90  \\
			DialogueCRN \cite{hu2021dialoguecrn} 	&   {6.57M}  &  {6.44s}  &  \textbf{71.47} 51.93 &  75.82 78.25 &  66.17 59.86 &  \textbf{78.53} 64.16 &  68.95 77.72 &  54.91  60.l8 \\
			SumAggGIN  \cite{sheng2020summarize}        	&  {14.28M}   &   {18.79s}                 & 56.74  54.22                                & 86.85  79.17                       & 62.95  65.32                       & 64.64  62.28                                & 76.21  78.43                       & 63.42  61.67                                              \\
			DisGCN  \cite{sun2021discourse} 	& {14.16M}    &  {17.85s}     &  71.17 {56.92}          &  68.65 76.47 &  66.63 57.41 & 74.26 54.35          &  74.54 76.47 & {51.14 59.28}   \\
			MM-DFN \cite{hu2022mm} 	&  {6.29M}   &   {6.27s}       & {40.17 42.22} & {74.27} 78.98 & \textbf{69.13 66.42} & {70.25} \textbf{69.97} & 76.99 75.56 & {\textbf{68.58}} 66.33\\

			M2FNet \cite{chudasama2022m2fnet}  	& {9.31M}    &  {10.37s}      & {65.92 60.00} & {79.18 82.11} & 65.80 {65.88} & {75.37 68.21} & {74.84 72.60} & {66.87 68.31} \\

			EmoCaps \cite{li2022emocaps} 	&  {13.41M}   &  {16.28s}         & 70.34 \textbf{72.86} & 77.39 82.45 & 64.27 65.10 & 71.79 69.14 & 84.50 73.90 &{63.94} 63.41  \\
			CT-Net \cite{lian2021ctnet}    	& {8.49M}    &   {9.06s}            & 47.97  51.36 & 78.01  79.94 & {69.08  65.82} & 72.98  67.21 & {85.35}  \textbf{78.74}         & 52.27  58.83   \\
			LR-GCN  \cite{ren2021lr}    	& {15.77M}    &    {21.06s}             & 54.24  55.51 & 81.67  79.14 & 59.13  63.84 & 69.47  {69.02} & 76.37  74.05 & 68.26  \textbf{68.91} \\  \hline
	\end{tabular}}
\end{table*}

\begin{table*}[!t]
	\renewcommand\arraystretch{1}
	\centering
	\caption{{On the MELD dataset, we counted the parameters, running time, and emotion recognition effects of different MCER algorithms on different emotion categories. The best result in each column is in bold.}}
	\label{tab:meld}
	\scalebox{0.72}{
		\setlength{\tabcolsep}{2.5pt}{
			\begin{tabular}{l|ccccccccc}
				\hline
				\multirow{3}{*}{Methods} & \multicolumn{7}{c}{MELD}                                                                            \\ \cline{2-10}
				& \multirow{2}{*}{{Parmas.}}  &  \multirow{2}{*}{{Running time}} & Neutral     & Surprise    & Fear     & Sadness    & Joy        & Disgust  & Anger       \\ \cline{4-10}
				&     &     & Acc.  F1    & Acc.  F1    & Acc.  F1 & Acc.  F1   & Acc.  F1   & Acc.  F1 & Acc.  F1      \\ \hline
				TextCNN \cite{kim2014convolutional}       &  \textbf{{0.34M}}   &  \textbf{{15.25s}}                 & 76.23  74.91  & 43.35  45.51  & 4.63  3.71 & 18.25  21.17 & 46.14  49.47 & 8.91  8.36 & 35.33  34.51  \\
				bc-LSTM  \cite{poria2017context}    &  {1.07M}   &    {20.29s}            & 78.45   73.84 & 46.82   47.71 & 3.84  5.46 & 22.47  25.19 & 51.61  {51.34} & 4.31  5.23 & 36.71  38.44  \\
				bc-LSTM+Att \cite{poria2017context} &  {1.66M}   &  {24.28s}  & 70.45 75.55 & 46.43 46.35 & 0.00 0.00 &21.77 16.27 & 49.30 50.72 & 0.00 0.00 & 41.77 40.71  \\
				A-DMN \cite{xing2020adapted}    & {6.25M}    &   {51.13s}            &  76.54 78.92       &  {56.24} 55.35        &   \textbf{8.22} 8.61     &   22.14  24.94       &  59.81 57.45      &   1.23 3.45     &   41.31  40.96       \\
				DialogueRNN \cite{majumder2019dialoguernn}    & {14.28M}    &   {141.31s}          & 72.12   73.54 & {54.42}  49.47  & 1.61  1.23 & 23.97  23.83 & 52.01  50.74 & 1.52  1.73 & 41.01  41.54  \\
				CT-Net \cite{lian2021ctnet}      &  {7.69M}   &  { 61.46s}          & 75.61   77.45       & 51.32  52.76        & 5.14  \textbf{10.09}     & 30.91  32.56       & 54.31  56.08       & {11.62  11.27}     & {42.51}  44.65        \\
				{DialogueCRN} \cite{hu2021dialoguecrn} &  {4.78M}   &  {40.35s}  & {70.91 75.73}            & {47.32 47.18}           & { 0.00 0.00}           & {{34.06} 13.29} & { 41.95 49.72}           & {0.00 0.00}           & {41.66 35.69}                   \\
				SumAggGIN  \cite{sheng2020summarize}    & {13.26M}    &  {122.69s}  &   {78.19 77.82} & {52.27 54.11}  & 2.17 2.31 & {35.79 36.43} & {54.15 55.07} & 4.05 {2.12} & {48.31 47.22}                    \\
				{DisGCN} \cite{sun2021discourse}  & {13.17M}    &  {108.71s}     & {70.84 76.67}            & {42.71 46.13}           & {1.17 1.55}           & {32.08 16.97}          & {50.03 50.17}           & {2.35 1.99}           & {38.25 39.97}                     \\
				MM-DFN \cite{hu2022mm}     & {5.33M}    &  {45.63s}           & {78.17 77.76}       & 52.15 50.69        & {0.00  0.00}     & 25.77  22.93       & {56.19  54.78}       & 0.00 0.00     & {48.31  47.82  }    \\
				M2FNet \cite{chudasama2022m2fnet} & {8.15M}    & {66.54s}   & {72.88 67.98} & \textbf{72.76} 58.66 & 5.57 3.45 & \textbf{50.09 47.03} & \textbf{68.49} 65.50 & \textbf{17.69 25.24} &  57.33 55.25\\
				EmoCaps \cite{li2022emocaps} &  {12.31M}   &  {101.77s}  & 75.24 77.12 & 63.57 \textbf{63.19} & {3.45 3.03} & {43.78 42.52} & 58.34 57.05 & 7.01 7.69 & \textbf{58.79 57.54} \\
				LR-GCN  \cite{ren2021lr}       &  {14.97M}   &   {147.38s}           & \textbf{81.51}   \textbf{80.83} & 55.42  {57.11}  & 0.00  0.00 & 36.36  36.96 & 62.21  \textbf{65.84} & 7.32  11.07 & {52.63}  54.74 \\ \hline
	\end{tabular}}}
\end{table*}

In addition, we also counted the emotion recognition effects of different MCER algorithms on different emotion categories. As shown in Table \ref{tab:IEMOCAP}, on the IEMOCAP data set, the performance effects of each algorithm on various emotions are consistent with the overall results introduced previously. The method based on context-free modeling has the worst effect, with the recognition effect on the ``happy'' emotion being less than 50\%. In comparison, most of the other three types of algorithms have exceeded 60\%, and some categories of emotions have exceeded 80\%. The performance of each algorithm on the MELD data set is shown in Table \ref{tab:meld}. The recognition effects of each algorithm on most categories of emotions are similar to those on the IEMOCAP data set. It is important to note that we found that all emotion recognition methods have poor performance in identifying ``fear'' and ``disgust'' emotions, and the accuracy of some algorithms is even 0\%. When we observe the distribution of the data set, we can find that the MELD data set has a serious data imbalance problem. This results in the model's very poor emotion recognition performance on minority classes.

{To comprehensively evaluate the practicality of different MCER models, we compare their parameter sizes and inference time in addition to their recognition performance. Tables \ref{tab:IEMOCAP} and \ref{tab:meld} report the number of parameters and average running time per inference on the IEMOCAP and MELD datasets, respectively. The results show that lighter models such as TextCNN and bc-LSTM possess significantly fewer parameters (0.47M and 1.28M, respectively) and exhibit faster inference times (0.96s and 2.16s). However, their performance across most emotion categories tends to be lower compared to more complex architectures. In contrast, recent transformer-based or graph-enhanced models, such as DialogueGCN, MM-DFN, M2ETNet, and especially LR-GCN, require substantially more parameters (up to 15.77M) and longer inference times (up to 21.06s for IEMOCAP and 147.38s for MELD), but consistently achieve better recognition performance. This observation highlights a common trade-off between model complexity and computational efficiency. While heavier models are better suited for applications where accuracy is the priority, lightweight models may be more favorable for real-time or resource-constrained deployment scenarios. In particular, LR-GCN achieves the best overall performance on both datasets, ranking highest in multiple emotion categories, albeit with the largest parameter count and slowest inference time. This suggests that while model complexity improves expressive power, there is a pressing need to explore model compression, pruning, and quantization to improve the feasibility of deploying such models in real-world applications.}

\section{Applications of Multi-modal Conversational Emotion Analysis}
\label{sec:sec7}
Emotion recognition is a method of applying natural language processing, machine learning, and deep learning techniques to multi-modal data such as text, video, and audio to identify and analyze the emotional state expressed in multi-modal data \cite{deng2021survey}. Therefore, analyzing and studying the problem of emotion recognition has broad application value in many practical application scenarios.

\subsection{Social Media Analysis}
Multi-modal conversational emotion recognition has many broad applications in social media analysis \cite{zhang2021real}. The most typical application is product improvement and innovation, that is, by analyzing user comments and feedback on social media, companies can understand users' preferences and dissatisfaction with products. This helps companies tweak product designs, improve functionality, and develop products that better meet user needs. Therefore, businesses can employ emotion analysis techniques to improve their products. In addition, emotion analysis can also help advertisers understand users' emotional attitudes towards advertisements, thereby optimizing advertisement content and strategies, and improving advertisement effectiveness. {For example, FaceReader \cite{li2024computer} can measure people’s emotional responses to different advertising creatives, providing valuable insights into the effectiveness of emotional appeals, humor, or shock value. With an 89\% recognition rate for static images and an 80\% recognition rate for animated expressions, FaceReader provides a reliable method for assessing the emotional impact of advertising imagery.}

\subsection{Public Opinion Analysis}
Multi-modal conversational emotion analysis also has a wide application value in opinion mining, which can help mine and analyze people's opinions and emotions expressed in text, video and audio \cite{tan2021speech}. {For example, emotion analysis of online public opinion on emergencies can better understand public emotion and conduct effective crisis management. ECR-BERT \cite{wan2024emotion} proposed a BERT-based model that integrates emotion-cognitive reasoning mechanisms, enabling more accurate emotional understanding in complex scenarios such as public opinion analysis during sudden events. Compared with the standard BERT model, ECR-BERT achieved absolute average accuracy improvements of 0.82\%, 1.74\%, 0.98\%, and 1.37\% across different datasets. This enhanced emotional recognition capability helps to more precisely capture public sentiment dynamics, providing valuable support for timely public opinion monitoring and effective crisis management.}


\subsection{Recommendations Systems}
Multi-modal conversational emotion analysis in recommender systems can help personalize recommendations more in line with users' emotions and preferences \cite{ding2022tsception}. For example, the recommendation system can recommend products that users are more interested in according to the emotional changes of consumers, and can perform emotion analysis on multi-modal data of user evaluations to realize real-time early warning and disposal of negative product evaluations. {For example, Agent4Rec \cite{zhang2024generative} displays four movies on each recommendation page, and the agent will decide whether to continue to the next recommendation page or exit the recommendation system based on their satisfaction. After the agent exits, the system will ask him to give a satisfaction score of the recommendation system, ranging from 1 to 10. Ratings above 3 are regarded as signals of liking. After the entire simulation is completed, the following multi-faceted indicators are collected: average viewing rate, average number of likes, average like ratio, average number of exit pages, and average user satisfaction score. The satisfaction score of random recommendations is 2.93, while the satisfaction score of the recommendation algorithm based on emotional preferences is 3.85. The experimental results prove the effectiveness of emotion recognition technology in the field of social media.}

\subsection{Medical Care}
Multi-modal conversational emotion analysis plays an important role in many aspects in the field of health care \cite{saganowski2022emotion}. It can help medical institutions and doctors better understand the current emotional state of patients, so as to give better treatment plans. {For example, by performing emotion analysis on unstructured data such as patients' medical records, consultation conversations, online consultation texts, or social media posts, the medical system can identify patients' potential emotional states, thereby providing supplementary psychological information for clinical diagnosis \cite{sun2024building}. In addition, in mental health assessments, emotion analysis can be used to detect early signs of depression tendencies or anxiety symptoms, which helps to achieve early screening and early intervention. In chronic disease management, the system can continuously monitor patients' emotional responses to treatment plans and help doctors dynamically adjust treatment strategies.}

\subsection{Financial Field Analysis}
In the field of financial analysis, emotion analysis can help financial practitioners and investors better understand the emotional state of the market and predict market trends, thereby helping investors make correct investment decisions \cite{gerczuk2021emonet}. {For example, the improved GPT model \cite{xingdesigning} combined with the emotion analysis module achieved an accuracy of 88.34\% in the financial emotion classification task, which is significantly better than the original GPT model (74.57\% accuracy) that did not use emotion information. The results show that emotion factors can provide the model with richer semantics and market tendency judgment basis, further verifying the actual effectiveness and research value of emotion analysis in financial text understanding and trend prediction.}

\subsection{Social Robot}
Multimodal conversational emotion recognition has many potential applications on social robots, which can enhance the capabilities of social robots and make them more intelligent and humane \cite{lee2022unboxing}. Social robots can use multimodal emotion recognition to sense the emotional state of the users they interact with. This includes identifying users’ facial expressions, voice emotions, text emotions, and other modal emotional signals \cite{laban2022informal}. The robot can then adjust its interaction to better meet the user's emotional needs, providing support, comfort or entertainment. In addition, social robots can use MCER to better understand users' needs and emotional states to provide personalized suggestions and assistance. {For example, emotion analysis can detect in real time the user's anxiety, frustration, or fatigue during the interaction process. The chatbot can then adjust its language response strategy and provide more soothing tone and suggestions, thereby significantly improving the user's trust and satisfaction. After the introduction of the emotion perception mechanism, the user's positive feedback rate on the chatbot has been significantly improved \cite{islam2024revolutionizing}.}

{\section{Privacy and Security of Multimodal Data}}

{With the widespread application of multi-modal conversational emotion recognition, privacy concerns have become increasingly critical \cite{yin2024primonitor}. Unlike unimodal data, multi-modal data often involves sensitive personal information spanning facial expressions, voice patterns, textual content, and physiological signals. These data types may reveal not only the user’s identity but also intimate emotional states, behavioral tendencies, and even mental health conditions. Consequently, the risk of personal emotional information leakage poses a significant challenge to the secure deployment of these technologies. To address these issues, privacy protection must be considered at every stage of system design and data processing. First, data anonymization techniques should be applied to remove or obfuscate identifying information, such as name, face, or unique voice features \cite{ye2024securereid, hanisch2025anonymization}. Second, data encryption techniques \cite{zhang2024heprune, rieyan2024advanced} should be used to ensure protection against unauthorized access to sensitive data, whether in storage, transmission, or computation. In a distributed environment, federated learning \cite{zhang2024enhancing} provides an effective framework in which model training is performed locally on the user's device and only encrypted model updates are shared with a central server, thereby protecting the privacy of the original data. Moreover, differential privacy \cite{xu2025privacy} can be introduced to inject calibrated noise into feature representations or model outputs, reducing the possibility of individual re-identification. In scenarios where multiple institutions or agents collaborate, secure multi-party computation and homomorphic encryption offer mechanisms for privacy-preserving joint model training or inference, albeit at the cost of computational efficiency \cite{xu2025privacy}. Finally, privacy-preserving representation learning is gaining traction, where adversarial training or disentangled learning techniques are used to suppress sensitive attributes (e.g., user identity) while preserving task-relevant emotional information \cite{feng2024robust}.}

\section{Research Challenges}
\label{sec:sec8}
Although deep learning technology has promoted the prosperity of MCER tasks, many scholars have proposed many state-of-the-art algorithms. However, building an accurate MCER model still faces challenges.

\subsection{Scarcity of Training Data}
Multi-modal conversation emotion recognition models require sufficient and comprehensive emotional samples as a basis to achieve accurate prediction or classification of emotions. The existing multi-modal benchmark data sets IEMOCAP, MELD, and SEMAINE have only 11098, 5810, and 394 utterances, respectively. Unfortunately, although we can easily collect large amounts of multi-modal conversation data from channels such as social media, the emotion labeling process is often expensive and time-consuming. In addition, the collected multi-modal data inevitably has problems such as ambiguous labels or multiple labels, which makes it a great challenge to obtain sufficient multi-modal labeled data, which in turn leads to the scarcity of multi-modal training data. Therefore, the scarcity of training data limits the effectiveness of current multi-modal conversational emotion recognition models.

\subsection{Data is Heterogeneous and Noisy}
MCER models need to fully eliminate heterogeneity and noise information between modalities to achieve accurate prediction or classification of emotions. Multi-modal data is naturally heterogeneous, and features of different modalities have huge differences in processing methods and representation forms. {Additionally, multimodal conversation data often contains a large amount of redundant or noisy information. The emotion is typically determined by a small amount of consistent key information, such as specific words in a sentence, a particular frequency band in speech, or a distinct expression in a video.} Even in some extreme cases, part of the modal information is basically unavailable under noise interference, such as ambiguous sentence expressions, noise in the speech, blocked expressions, etc. Therefore, the heterogeneity and noise of data limit the effectiveness of current multi-modal conversational emotion recognition models.

\subsection{Unbalanced Data Distribution}
Multi-modal dialogue data samples have serious imbalance problems, and the unbiased learning of the model is seriously interfered with. The multi-modal conversation emotion recognition model is based on cross-modal feature fusion, driven by emotion category sample data, and is easily affected by the number of emotion category samples. However, multi-modal conversation emotion data naturally suffers from the problem of category sample imbalance. A few emotion category samples account for a larger proportion, while most emotion category samples account for a small proportion. For example, in the MELD data set, the ``fear'' emotion only accounts for 1.91\% of the total samples, and the ``disgust'' emotion only accounts for 2.61\% of the total samples. A similar sample distribution also exists on the benchmark data set SEMAINE. Small samples are difficult to drive unbiased learning of the model, which seriously affects the model's prediction accuracy for small sample emotional categories. Therefore, the unbalanced sample distribution limits the effectiveness of current multi-modal conversational emotion recognition models.

\subsection{Consistent Semantic Association}
{Multimodal conversation emotion recognition requires the model to learn the consistent semantics across modalities to filter out noise and eliminate heterogeneity. This is essential for building an accurate multimodal emotion recognition model.} {However, the consistent semantic association in multimodal conversation is more complex. It is not only related to the multimodal context but also influenced by factors such as the conversation scene, the speaker's emotional inertia, and their responses.} In addition, multi-modal data are heterogeneous, each modality has differentiated representation and distribution characteristics in space, and some consistent semantic associations are hidden in the feature distribution space between modalities. Therefore, efficiently performing consistent semantic association is the primary issue that needs to be considered at the model level.

\subsection{Complementary Semantic Capture}
Multi-modal conversation emotion recognition models need to establish accurate and consistent semantic associations and capture complementary semantic features between modalities, which can expand the emotional representation capabilities of a single modality. However, unlike consistency semantics, complementary semantics represent differences between modalities, and this difference may contain noise components. Therefore, consistency semantics and complementarity semantics are a pair of game entities, and how to balance the relationship between them is another issue that needs to be considered at the model level.

\subsection{Multi-model Collaboration}
Multi-model collaboration is the third challenge faced at the model level in building accurate multi-modal conversation emotion recognition models. Multi-modal conversation emotion recognition often requires the collaboration of multiple models to complete tasks, such as feature extraction models and feature fusion models. However, existing methods often perform task collaboration from the data level and ignore the collaborative relationship between models. Therefore, in order to achieve ideal synergistic results, not only the respective characteristics of the modes and their interrelationships need to be considered, but also the synergistic relationships between models need to be considered.

\section{Future Work}
\label{sec:sec9}
\subsection{Multi-modal Conversation Data Generation}
Multi-modal conversational emotion recognition models require sufficient and comprehensive emotional samples as a basis. When sample data is scarce, training multi-modal conversation emotion recognition models without causing overfitting or underfitting problems is extremely challenging. However, the sample size of existing benchmark data sets is relatively small, and there is a common problem of data scarcity. Multimodal dialogue data generation can effectively alleviate this problem. However, the distribution of multi-modal conversation data is more complex, and traditional single-modal data generation or cross-modal data generation models cannot meet the requirements. Therefore, there is an urgent need to solve the problem of collaborative generation of multi-modal conversation data.


{To solve the problem of ensuring strong correlation and synergy among modalities in Multi-modal Conversation Data Generation, researchers have designed advanced generative frameworks that explicitly model the cross-modal dependencies during the generation process. One representative approach leverages variational autoencoders (VAEs), which map different modalities into a shared latent semantic space, allowing the model to capture deep inter-modal relationships and generate coherent multi-modal conversational data through joint reconstruction. This shared representation ensures that generated text, audio, and visual signals are semantically aligned and contextually consistent. In addition, GAN-based frameworks introduce modality-specific discriminators alongside a joint discriminator to constrain both the individual quality and overall coherence of generated modalities. By adversarially optimizing the generator to produce modality-consistent outputs, these methods effectively enhance cross-modal correlation in generated conversations. Recently, diffusion models have shown strong potential for multi-modal data generation by modeling the complex joint distribution of multiple modalities through iterative denoising steps. Diffusion-based methods \cite{ho2020denoising} can incorporate cross-modal conditional signals at each generation step, ensuring that the evolving text, audio, and visual outputs remain temporally synchronized and semantically coupled.}

\subsection{Multi-modal Feature Deep Fusion}
Multi-modal feature fusion is crucial to the MCER task. The fused feature vector can represent the consistent semantics and complementary information between modalities. However, many different information interactions exist between multi-modalities, and many consistent or complementary features are hidden in multiple time series or local spatial correlations. Since multi-modal conversation data is heterogeneous and contains noise, there are significant differences in the temporal period and spatial distribution of different modal features, and the spatiotemporal importance between modalities is dynamic. Currently, few works consider this difference, and more efforts are still needed for deep fusion of multi-modal features.

{To solve the above problems, on the one hand, the deformable temporal convolution can be used to allow each modality to dynamically sample the most relevant time step based on its own features. Then, the locality-aware attention is utilized to focus on the strong correlation of local areas in space. The time period misalignment and local information loss caused by heterogeneity are solved before fusion. On the other hand, to achieve cross-modal alignment, a dynamic weighted alignment mechanism can be introduced to calculate the dynamic weights of consistency and complementarity between modalities for each moment or local spatial area. Through the cross-modal attention module, the consistency score and complementarity score of each modality at the current spatiotemporal point are calculated. Then, the dynamic gating mechanism is used to adaptively adjust the contribution weight of each modal feature during fusion according to the score to avoid information redundancy or key feature loss caused by static weighting.}

\subsection{Unbiased Emotional Learning}
Many benchmark datasets in the field of multi-modal conversational emotion recognition suffer from serious sample category imbalance, that is, the minority emotion category contains a large amount of data, while the majority category emotion only contains a small amount of data. In the case of unbalanced data, the existing models tend to be biased towards fitting the minority emotion with a large amount of data, and the learning is insufficient on the majority emotion with a small amount of data, which leads to the model being in a small sample emotion category, resulting in the recognition accuracy is poor. Thus, the small-sample problem in multi-modal dialogue emotion recognition urgently requires further research.

{To effectively solve the problem of small samples in MCER, we can start from three dimensions: enhancing samples, optimizing model structure, and adjusting training strategies, and build an integrated framework of data augmentation, prototype modeling, and category balance training. Specifically, at the data augmentation level, the emotional sentences of small sample categories are combined with context to reconstruct new samples, and the diversity of small sample categories is improved while ensuring the rationality of the context. At the prototype modeling level, a multimodal prototype center is constructed for each emotion category, and the distance between samples and prototypes of the same type is shortened by contrastive loss, and the distance from prototypes of other types is pushed away to alleviate the problem of inter-class imbalance caused by differences in sample size. At the category balance training level, the category balance focal loss or label distribution-aware margin loss is used to dynamically adjust the loss penalty items of each category. Without changing the overall training process, the model's attention to small sample categories is improved.}

\subsection{Incomplete Multi-modal Conversation Emotion Recognition}
Each modality is not always available in real-world scenarios, which can lead to modal incompleteness problems. For example, the voice contains much noise, the expression is blocked, the light is dim, etc. At this moment, some modal information becomes unavailable due to noise interference. Modal integrity requirements reduce the applicability of multi-modal conversation emotion recognition methods. Therefore, cross-modal content recovery methods based on deep learning should continue to be developed to achieve multi-modal conversation emotion recognition in missing modalities.

{To solve the common problem of missing modalities in MCER, researchers have proposed a variety of modal restoration methods in recent years, aiming to maintain the discriminative ability of the model when some modalities are unavailable. A common type of method is based on autoencoders or variational autoencoders (VAEs), which reconstruct the representation of the missing modality using the available modalities by learning the mapping relationship between different modalities. Another type of method uses generative adversarial networks (GANs) to constrain the generated features by introducing discriminators to improve the authenticity and diversity of modality completion. There are also studies that use knowledge distillation strategies to guide student models to maintain performance in modality-incomplete scenarios with the help of teacher models trained under complete modalities. Recently, the development of generative models has also provided stronger modeling capabilities for modality restoration. Diffusion models have strong distribution modeling and step-by-step optimization capabilities by modeling the reverse generation process from noise to data. They perform well in conditional generation tasks and are suitable for high-quality restoration of missing modalities. Meanwhile, flow-based models accurately model the joint distribution of modalities through reversible transformations and support conditional sampling to achieve missing modality completion.}

\subsection{Zero-shot Multi-modal Conversation Emotion Recognition}
Affected by factors such as the complexity of emotions and the high cost of labeling, it is difficult to fully label some emotional samples. Furthermore, with the rapidly growing personal emotion annotation space, real-world emotion recognition systems may frequently encounter unseen emotion labels. Therefore, improving the generalization performance of emotion recognition models is an issue that needs to be considered. Deep methods utilizing zero-shot learning are expected to achieve better multi-modal dialogue emotion recognition.

{In recent years, large-scale pretrained models, especially cross-modal models such as CLIP \cite{radford2021learning}, Flamingo \cite{alayrac2022flamingo}, GPT-4V \cite{yang2023dawn}, etc., have brought a new paradigm for zero-shot multimodal emotion recognition. These models usually have strong multimodal alignment capabilities and natural language understanding capabilities, and can achieve category expansion through contrastive learning or generative modeling of text-image or text-audio. For example, CLIP builds a shared embedding space through image-text contrastive learning, which can convert emotion labels into natural language descriptions to achieve open-class emotion recognition. Audio-text models such as Whisper \cite{cao2012whisper} and AudioCLIP \cite{guzhov2022audioclip} are also used to map speech emotion embeddings to language space to achieve zero-shot emotion transfer. In addition, the generative capabilities of multimodal large models (MLLMs) \cite{liu2023llava} can transform emotion recognition tasks into natural language generation or question-answering problems, and complete the understanding and prediction of new emotion categories under unsupervised or weak supervision. Combined with strategies such as prompt engineering, instruction tuning, or emotionally-informed prompts, the model can generalize to new emotion categories with only label descriptions.}

\subsection{Multi-modal Conversation Multi-label Emotion Recognition}
In multi-modal conversation scenarios, existing emotion recognition models usually use a single-label supervised learning. Due to the ambiguity of emotions, emotion recognition in real life is often a multi-label task. The single-label requirement greatly limits the application scenarios of multi-modal conversation emotion recognition. Therefore, the multi-label emotion recognition problem in multi-modal conversation scenarios should be considered in future work.

{To address the multi-label expression problem in MCER, researchers have proposed a range of targeted solutions to capture the complexity and coexistence of emotions within conversations. A common approach is to replace traditional Softmax classification with Sigmoid activation, enabling the model to independently predict the probability of each emotional label and naturally support the coexistence of multiple emotions such as sadness and anger. In addition, label dependency modeling is widely adopted, where the statistical co-occurrence patterns of emotions are explicitly captured using techniques such as label graphs or graph neural networks (GNNs), ensuring semantic consistency and reducing contradictory label outputs. To further improve robustness in real-world noisy environments, uncertainty-aware mechanisms have been integrated into multi-label frameworks, allowing the model to dynamically adjust label confidence based on the reliability of different modalities. Moreover, some studies introduce auxiliary tasks, such as emotion intensity regression or label quantity estimation, to provide richer supervision signals and enhance the model's ability to represent complex emotional states. Recently, researchers have also explored label-specific attention mechanisms, dynamically modulating the contribution of each modality for different emotional labels, which effectively improves the multi-label recognition accuracy under modality imbalance or incomplete scenarios.}

{\subsection{Multi-modal Emotion Recognition in Dynamic Dialogue Scenarios}}

{Real-world conversations are inherently dynamic, with speaker roles changing depending on the context and interpersonal relationships evolving over time. However, most current multi-modal conversational emotion recognition (MCER) methods rely on static modeling frameworks that fail to capture these real-time dynamics. To address this gap, future research should focus on developing models that can adapt to such temporal and structural variability in conversation.}

{One promising direction is the use of time-series modeling techniques. Methods such as Temporal Convolutional Networks (TCN) \cite{hussain2024big, sheng2024residual} and Hierarchical Recurrent Neural Networks (HRNN) \cite{sahin2024nonlinear, zhou2024design} can be employed to model frame-level and utterance-level emotional fluctuations across a dialogue sequence. These models can maintain temporal dependencies and detect sudden or gradual emotional transitions by capturing long-range contextual signals. For example, by feeding sequential utterance embeddings into a TCN layer with causal convolutions, the model can learn how earlier statements influence emotional progression. Another critical aspect is the dynamic modeling of speaker interactions. Dynamic Graph Neural Networks (Dynamic GNNs) \cite{yang2024emotion, wang2024dynamic, zhang2024tt} provide a powerful framework to represent evolving dialogue structures. Here, each node in the graph represents a speaker utterance or a speaker entity, and edges represent context-aware relationships (e.g., speaker interactions, turn-taking). Unlike static GCNs, dynamic GNNs update node and edge embeddings over time based on incoming utterances and relational changes. For instance, Temporal Graph Networks (TGNs) and EvolveGCN can learn temporal node representations by integrating recurrent modules that evolve the graph state as the dialogue progresses. These approaches allow the system to adjust for speaker role shifts and emotional influence propagation between participants. Additionally, speaker-role modeling mechanisms can be introduced by using speaker-aware encoders that condition emotion predictions on dynamic role embeddings. For example, using attention modules that incorporate speaker identity, conversational history, and position in the dialogue tree can help model role-specific emotional behaviors. Coupled with adaptive memory modules (e.g., Transformer-based memory networks), the system can retain evolving emotional cues and speaker traits across multiple turns. Finally, hybrid systems that combine reinforcement learning for emotion trend tracking and graph-based relational reasoning can adaptively adjust prediction strategies based on dialogue context evolution. This provides a more robust mechanism for handling emotional ambiguity and inter-speaker dynamics in realistic, multi-turn conversations.}

{\subsection{Lightweight and Efficient Multimodal Conversational Emotion Recognition}}

{Although most existing MCER models have demonstrated strong emotion recognition performance, their computational complexity and resource requirements hinder deployment in practical applications, especially on resource-constrained platforms such as mobile devices, wearable devices, and embedded systems. In real-world scenarios, real-time emotion recognition is essential for applications such as mobile assistants, edge-based healthcare monitoring, and socially interactive robots. Therefore, designing lightweight and efficient MCER models is critical for practical adoption.}

{To achieve real-time emotion recognition, the deployment efficiency of the model on resource-constrained devices (such as mobile terminals and embedded systems) must be solved. One of the key paths is to compress and optimize the emotion recognition model so that it can significantly reduce the computational overhead while maintaining high recognition accuracy. First, model compression can transfer the predictive ability of a large model (i.e., teacher model) trained on a high-performance platform to a smaller model (student model) with a more compact structure through methods such as knowledge distillation. During the distillation process, the student model not only learns the original labels of the training data, but also learns the "soft labels" output by the teacher model, that is, the probability distribution of each emotion category. This method allows the small model to learn richer feature representations. The compressed model greatly reduces the number of parameters and inference delay without sacrificing too much accuracy. Secondly, pruning technology mainly evaluates the importance of certain neurons or channels in the network to the final output and prunes the less influential parts. Pruning is usually divided into two categories: structured pruning and unstructured pruning. Structured pruning can remove the entire convolution kernel, channel or layer. This method is more hardware-friendly and easy to accelerate and parallelize. Unstructured pruning performs sparse processing at the weight level. Although it has more detailed control over precision loss, it requires special sparse matrix acceleration support in actual deployment. Through multiple rounds of pruning and fine-tuning, the model complexity can be further reduced and the inference speed can be improved. Finally, the quantization method converts the original floating-point model parameters and intermediate activation values into a lower bit-width representation, such as compressing from 32-bit floating points to 8-bit integers (INT8) or even lower bit widths. Quantization can significantly reduce the storage requirements and memory bandwidth consumption of the model, while fully utilizing the integer operation acceleration capabilities of the hardware on many mobile and edge computing devices. In order to avoid the decline in recognition effect due to reduced accuracy, strategies such as quantization-aware training (QAT) or post-training quantization (PTQ) are usually used to fine-tune the weights and activation distributions to maintain the accuracy of emotion recognition.}

\section{Conclusion}
This paper reviews the latest research results in the field of multi-modal conversational emotion recognition. To allow readers to implement emotion recognition tasks better, we have collected popular data sets in this field and given relevant download links. Since text, video, and audio are unstructured data that cannot be directly input into a computer for computation, we summarize some publicly available feature extraction methods. We divide emotion recognition methods into four categories, i.e., context-free modeling, sequential context modeling, distinguishing speaker modeling, and speaker relationship modeling. This paper further discusses the challenges faced by existing methods and future research directions. According to the review of existing work, it is found that multi-modal emotion recognition mainly improves the effect of emotion recognition by modeling intra-modal and inter-modal complementary semantic information. We hope this review can shed some light on developments in this field.

\section{Acknowledgments}
This work is supported by the National Natural Science Foundation of China (Grant No. 62372478, 62472165), the YueLuShan Center Industrial Innovation (2026YCII0126), the Research Foundation of Education Bureau of Hunan Province of China (Grant No. 22B0275), the Hunan Provincial Natural Science Foundation General Project (Grant No. 2025JJ50380), and the Hunan Provincial Natural Science Foundation Youth Project (Grant No. 2025JJ60420).

\bibliographystyle{ACM-Reference-Format}
\bibliography{refs}


\end{document}